\pdfoutput=1

\documentclass[11pt]{article}

\PassOptionsToPackage{table}{xcolor}

\usepackage[final]{acl}

\usepackage{times}
\usepackage{latexsym}
\usepackage[T1]{fontenc}
\usepackage[utf8]{inputenc}
\usepackage{microtype}
\usepackage{inconsolata}
\usepackage{graphicx}
\usepackage{booktabs}
\usepackage{soul}
\usepackage{empheq}
\usepackage{wrapfig}
\usepackage{enumitem}
\setlist[itemize]{leftmargin=*}
\setlist[enumerate]{leftmargin=*}
\usepackage{caption}
\usepackage{subcaption}
\usepackage{cleveref}
\usepackage{amsmath, amssymb, amsfonts}
\usepackage{hyphenat}
\usepackage[most]{tcolorbox}
\usepackage{listings}
\definecolor{pos}{HTML}{C6EFCE}
\definecolor{neg}{HTML}{FFC7CE}
\definecolor{headerbg}{HTML}{D3D3D3}
\definecolor{groupbg}{HTML}{F0F0F0}
\definecolor{hlyellow}{HTML}{FCEFCB}
\definecolor{amethyst}{rgb}{0.6, 0.4, 0.8}

\title{Prior Prompt Engineering for Reinforcement Fine-Tuning}

\author{
  \textbf{Pittawat Taveekitworachai\textsuperscript{1}},
  \textbf{Potsawee Manakul\textsuperscript{1}},\\
  \textbf{Sarana Nutanong\textsuperscript{2}},
  \textbf{Kunat Pipatanakul\textsuperscript{1}}
\\
\\
  \textsuperscript{1}SCB 10X R\&D,\\ SCB 10X, SCBX Group, Thailand\\
  \textsuperscript{2}School of Information Science and Technology,\\ Vidyasirimedhi Institute of Science and Technology, Thailand
\\
  \small{
    \href{mailto:pittawat@scb10x.com}{pittawat@scb10x.com},
    \href{mailto:potsawee@scb10x.com}{potsawee@scb10x.com},
    \href{mailto:snutanon@vistec.ac.th}{snutanon@vistec.ac.th},
    \href{mailto:kunat@scb10x.com}{kunat@scb10x.com}
  }
}

\begin{document}
\maketitle

\begin{abstract}

This paper investigates prior prompt engineering (pPE) in the context of reinforcement fine-tuning (RFT), where language models (LMs) are incentivized to exhibit behaviors that maximize performance through reward signals. While existing RFT research has primarily focused on algorithms, reward shaping, and data curation, the design of the prior prompt--the instructions prepended to queries during training to elicit behaviors such as step-by-step reasoning--remains underexplored. We investigate whether different pPE approaches can guide LMs to internalize distinct behaviors after RFT. Inspired by inference-time prompt engineering (iPE), we translate five representative iPE strategies--reasoning, planning, code-based reasoning, knowledge recall, and null-example utilization--into corresponding pPE approaches. We experiment with Qwen2.5-7B using each of the pPE approaches, then evaluate performance on in-domain and out-of-domain benchmarks (e.g., AIME2024, HumanEval+, and GPQA-Diamond). Our results show that all pPE-trained models surpass their iPE-prompted counterparts, with the null-example pPE approach achieving the largest average performance gain and the highest improvement on AIME2024 and GPQA-Diamond, surpassing the commonly used reasoning approach. Furthermore, by adapting a behavior-classification framework, we demonstrate that different pPE strategies instill distinct behavioral styles in the resulting models. These findings position pPE as a powerful yet understudied axis for RFT.

\end{abstract}

\section{Introduction}\label{sec:intro}

\begin{figure*}[htbp]
\begin{center}
\includegraphics[width=\linewidth]{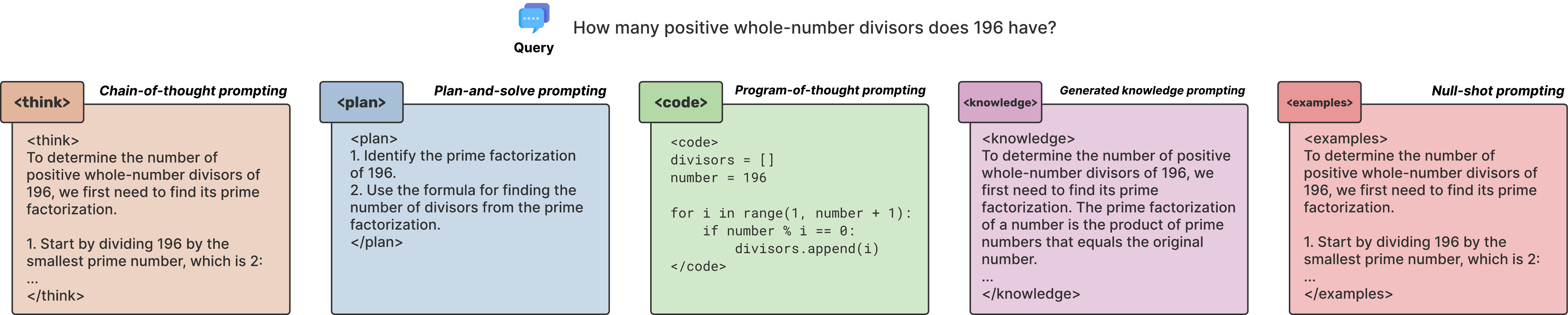}
\end{center}
\caption{Five generated responses from five distinct models post-RFT with different pPE approaches--\texttt{<think>}, \texttt{<plan>}, \texttt{<code>}, \texttt{<examples>}, and \texttt{<knowledge>}. Each pPE approach is inspired by a corresponding iPE paradigm: chain-of-thought, plan-and-solve, program-of-thought, null-shot, and generated knowledge prompting, respectively.}
\label{fig:hero}
\end{figure*}

Recent advancements in reasoning models mark a significant step forward in improving language model (LM) performance by allocating additional compute budget at test time. A common approach to developing such models is reinforcement fine-tuning (RFT), which incentivizes an LM to perform extended reasoning during inference by using reward signals--based on the correctness of generated answers--during training. Current studies have explored various components of the RFT pipeline, including objective functions and training algorithms \citep{liu2025understandingr1zeroliketrainingcritical,yu2025dapoopensourcellmreinforcement,yeo2025demystifyinglongchainofthoughtreasoning,yue2025vapoefficientreliablereinforcement}, data domains and curricula \citep{xie2025logicrlunleashingllmreasoning,wei2025swerladvancingllmreasoning,su2025crossingrewardbridgeexpanding,hu2025openreasonerzeroopensourceapproach}, reward functions and shaping \citep{yeo2025demystifyinglongchainofthoughtreasoning,su2025crossingrewardbridgeexpanding,hu2025openreasonerzeroopensourceapproach}, and the influence of inherent behaviors across different LM families and model sizes \cite{liu2025understandingr1zeroliketrainingcritical,zeng2025simplerlzooinvestigatingtamingzero,gandhi2025cognitivebehaviorsenableselfimproving}. However, despite these improvements for various components of the RFT pipeline, one critical aspect remains understudied: \emph{the design of the prompt}.

\begin{figure}[htbp]
\centering
\footnotesize
\begin{tcolorbox}[colback=gray!5, colframe=amethyst!75!black, breakable]
{\sethlcolor{hlyellow}\hl{A conversation between User and Assistant. The user asks a question, and the Assistant solves it. The assistant first thinks about the reasoning process in the mind and then provides the user with the answer. The reasoning process and answer are enclosed within $<$think$>$ $<$/think$>$ and $<$answer$>$ $<$/answer$>$ tags, respectively, i.e., $<$think$>$ reasoning process here $<$/think$>$$<$answer$>$ answer here $<$/answer$>$.}} User: Let the circles $k_{1}$ and $k_{2}$ intersect at two distinct points $A$ and $B$, and let $t$ be a common tangent of $k_{1}$ and $k_{2}$, that touches $k_{1}$ and $k_{2}$ at $M$ and $N$, respectively. If $t \perp A M$ and $M N=2 A M$, evaluate $\angle N M B$. Assistant:
\end{tcolorbox}
\caption{The prompt used during RFT by \citet{deepseekai2025deepseekr1incentivizingreasoningcapability}. The prior prompt is highlighted in {\sethlcolor{hlyellow}\hl{yellow}}. Non-highlighted content is task content.}
\label{fig:prior_deepseek_highlighted}
\end{figure}

To scope our study, we separate a prompt used during RFT into two main components: the instruction and the task content (see \Cref{fig:prior_deepseek_highlighted}). The instruction guides the model to exhibit desired behaviors (e.g., step-by-step reasoning). We refer to this section as the \textbf{prior prompt}, which is the main focus of this study. Examples of prior prompts from existing work are provided in \Cref{sec:prior_prompt_examples}. While some studies briefly note the role of prior prompts in training stability and performance \citep{xie2025logicrlunleashingllmreasoning,zeng2025simplerlzooinvestigatingtamingzero}, there has been little systematic investigation into how different prior prompting approaches during RFT shape model behaviors. This study therefore centers on the following question: \emph{Can different prior prompt engineering approaches guide language models to internalize distinct behaviors during RFT?}

The breadth of prompt engineering, which we define in this paper as \textbf{inference-time prompt engineering} (iPE) to distinguish it from prompt engineering during training, demonstrates its effectiveness in eliciting diverse behaviors (i.e., generation patterns) from LMs \citep{kojima2022large}, ultimately leading to varying performance outcomes. For instance, chain-of-thought prompting (CoT) \citep{NEURIPS2022_9d560961} elicits step-by-step reasoning before producing a final answer; plan-and-solve prompting (PS) \citep{wang-etal-2023-plan} first generates a high-level plan before problem solving; and program-of-thought prompting (PoT) \citep{chen2023program} induces code-based reasoning. These examples illustrate that different iPE approaches not only elicit distinct behaviors (e.g., reasoning, planning, coding) but also lead to varied performance results.

Inspired by iPE, we introduce the term \textbf{prior prompt engineering} (pPE) to denote approaches for modifying the prior prompt in RFT. Just as iPE guides behavior during \emph{inference}, we conjecture that pPE can shape model behavior during \emph{training}. By combining the varied elicitation induced by pPE with RFT’s incentivization mechanism, the resulting models may exhibit diverse behaviors and achieve different levels of performance impact.

In this paper, we study the effects of various pPE approaches on model behaviors and performance impact after RFT. We select five representative iPE approaches based on their distinct elicited behaviors--reasoning, planning, coding, knowledge recall, and null example utilization--and translate them into corresponding pPE approaches. We employ these five pPE approaches to train Qwen2.5 7B into five distinct models with RFT using math-only training data. We then compare each RFT-trained model to its corresponding iPE-only baseline.

We evaluate our models using both quantitative and qualitative methods. Quantitatively, we measure performance on mathematical reasoning, coding, and question-answering benchmarks (e.g., AIME2024, GPQA Diamond, and HumanEval+) to assess impact on in-domain and out-of-domain tasks. Qualitatively, we employ a modified behavior-classification framework from \citet{gandhi2025cognitivebehaviorsenableselfimproving} to quantify differences in model behaviors. To test generalization, we replicate our experiments at smaller scales on Qwen2.5 3B, Qwen2.5 Coder 7B, and Llama 3.1 8B.

We find that all pPE-trained models surpass their corresponding iPE-only baselines. Among pPE approaches, the null-example utilization approach--which exhibits behavioral similarities to the reasoning approach--achieves the largest improvement on GPQA Diamond and the highest average performance gain across tasks. \Cref{fig:hero} illustrates the five models trained with different pPE strategies, each demonstrating distinct behavioral styles and indicating that pPE can incentivize diverse behaviors. Our contributions are as follows:

\begin{itemize}
    \item We introduce the concepts of \textbf{prior prompt} and \textbf{prior prompt engineering} (pPE) as critical yet previously understudied aspects of RFT.
    \item \textbf{We demonstrate that different pPE strategies elicit distinct behaviors}, including variations in performance impact, response structure, verbosity, and behavior types.
    \item We propose \textbf{an updated systematic behavior classification approach} to quantify both cognitive and elicited behaviors, revealing how different pPE approaches shape model behavior.
\end{itemize}

\section{Prior Prompt Engineering for Reinforcement Fine-Tuning}\label{sec:ppe_rft}

\begin{figure*}[htbp]
  \centering
  \includegraphics[width=\linewidth]{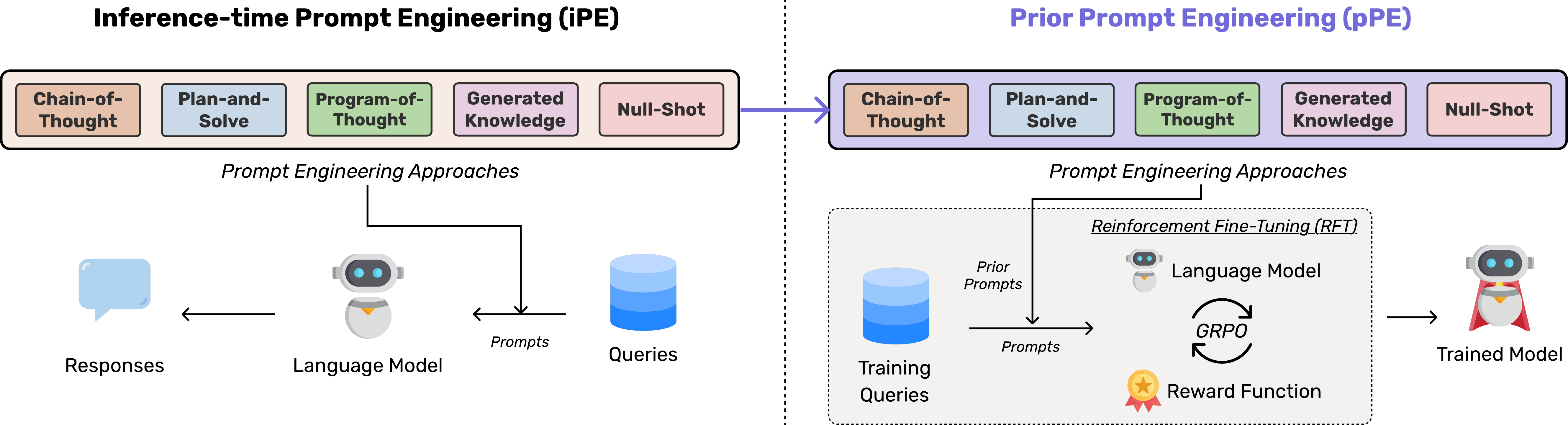}
  \caption{\textit{Left:} iPE approaches are applied to a prompt during inference, before inputting it into an LM, to elicit desired behaviors in the response. \textit{Right:} pPE approaches are translated from iPE approaches and applied to the prior prompt to elicit desired behaviors during training.}
  \label{fig:approach}
\end{figure*}

Our main question in this study is whether different pPE approaches can lead an LM to internalize distinct behavioral styles after RFT. If different pPE approaches indeed yield different behaviors, this could provide a simple means--by only changing the prior prompt--to train models for specialized behaviors beyond reasoning (e.g., plan generation, code-based reasoning, or knowledge generation). To answer this question, we select five representative iPE approaches and translate them into pPE approaches. We then apply a standard RFT setup to train five distinct models, differing only in their pPE approach and format reward (see \Cref{sec:exp_setup}). The overall process and distinctions between iPE and pPE are depicted in \Cref{fig:approach}.

We evaluate each model quantitatively and qualitatively to assess performance changes and behavioral differences. Quantitative evaluation uses established benchmarks for mathematical reasoning, coding, and question answering. For qualitative evaluation, we adapt the framework of \citet{gandhi2025cognitivebehaviorsenableselfimproving} to classify each post-RFT model’s behavior into one of four cognitive categories and five pPE-specific categories. We also apply these evaluations to the base model at inference time with different iPE approaches, to further compare iPE and pPE.

To explore the impact of pPE approaches on prior prompts, we select five representative iPE approaches based on their differences in behavioral elicitation when used to prompt an LM:

\begin{enumerate}
  \item \textbf{Reasoning:} Chain-of-thought prompting (CoT) \citep{NEURIPS2022_9d560961} elicits an LM to generate step-by-step reasoning before producing a final answer. This iPE approach is mapped to \texttt{<think>} in pPE and is the most commonly used in RFT studies, resulting in reasoning models \citep{deepseekai2025deepseekr1incentivizingreasoningcapability, xie2025logicrlunleashingllmreasoning}. This serves as our baseline for comparison.
  \item \textbf{Planning:} Plan-and-solve prompting (PS) \citep{wang-etal-2023-plan} elicits the model to first generate a plan (e.g., numbered steps) and then execute that plan, yielding improvements over standard CoT. The planning approach is mapped to \texttt{<plan>} in pPE. We expect the post-RFT model to generate a plan before providing an answer.
  \item \textbf{Code-based reasoning:} Program-of-thought prompting (PoT) \citep{chen2023program} elicits structured reasoning through code by asking a model to generate relevant code for problem solving. PoT has shown strong performance on math and logic tasks, especially with code-pretrained models such as CodeLlama \citep{roziere2024codellamaopenfoundation}, Qwen2.5-Coder \citep{hui2024qwen25codertechnicalreport}, and StarCoder 2 \citep{lozhkov2024starcoder2stackv2}. This iPE approach is mapped to \texttt{<code>} in pPE. We expect the post-RFT model to generate code and comments that solve the given task.
  \item \textbf{Knowledge recall:} Generated knowledge prompting \citep{liu-etal-2022-generated} asks the model to recall or synthesize relevant knowledge before answering, simulating a form of self-retrieval and improving performance on commonsense benchmarks. This approach is mapped to \texttt{<knowledge>} in pPE. We expect the post-RFT model to recall definitions, theorems, or formulas before proceeding to a final answer.
  \item \textbf{Null-example utilization:} Null-shot prompting \citep{taveekitworachai-etal-2024-null} prompts the model to utilize non-existent in-context examples relevant to the question, exploiting inductive biases without providing real demonstrations. It maps to \texttt{<examples>} in pPE, and we expect the post-RFT model to generate or reference illustrative examples relevant to a query.
\end{enumerate}

With these five distinct pPE approaches for eliciting different behaviors in LMs during RFT, we expect not only differences in performance impact and post-RFT behaviors but also in training dynamics, such as average response length or per-step reward trajectories.

\section{Experimental Setup}\label{sec:exp_setup}

\subsection{Prior Prompts}
To construct our prior prompts, we adapt the template of \citet{xie2025logicrlunleashingllmreasoning}. For each iPE approach, we modify the instruction in the template (e.g., ``plan,'' ``recall relevant knowledge,'' ``write required code'') and update the corresponding tag \texttt{<x></x>} as described in \Cref{sec:ppe_rft}. The \texttt{<think></think>} pPE approach thus is the standard RFT setup. These same templates are also used when evaluating iPE-prompted models. The complete prior prompt templates are provided in \Cref{sec:prior_prompts}.

\subsection{Training}\label{sec:training}

We follow a standard RFT setup similar to \citet{deepseekai2025deepseekr1incentivizingreasoningcapability}. Specifically, we use Group Relative Policy Optimization (GRPO) with a pretrained base LM. Our training stack is OpenRLHF v0.6.4 \citep{hu2024openrlhfeasytousescalablehighperformance} for policy optimization and vLLM v0.8.2 \citep{10.1145/3600006.3613165} for rollout generation. We train using prompts from the STILLv3 dataset \citep{chen2025empiricalstudyelicitingimproving}, which contains approximately 30K mathematical problems and is used to train a reasoning model. We note that the use of math-only training datasets is common in the existing literature \citep{liu2025understandingr1zeroliketrainingcritical,yeo2025demystifyinglongchainofthoughtreasoning,yu2025dapoopensourcellmreinforcement}. In addition, math-only training datasets provides a simplicity in verifiable reward design, i.e., value equivalent checking between a generated answer and the ground truth, unlike other domains, which inconclusive in implementation standards of the reward function. Additional details and hyperparameters are listed in \Cref{sec:training_script}.

Our reward function comprises two equally weighted components (summing to 1.0): (1) \emph{accuracy}, which assesses whether the model produces the correct final answer; and (2) \emph{format}, which assesses whether the model's output follows the expected format--\texttt{<x></x>} followed by \texttt{<answer></answer>}, where \texttt{x} is one of \{\texttt{think}, \texttt{plan}, \texttt{code}, \texttt{knowledge}, \texttt{examples}\}. The expected format is updated dynamically to match the pPE approach. Additional details on the reward function are available in \Cref{sec:reward_design}.

For our main experiments, we use Qwen2.5-7B \citep{qwen2025qwen25technicalreport} as the base model. All five pPE variants are trained with the same settings, differing only in the pPE approaches. We also use Qwen2.5-7B, prompted at inference with each iPE approach, as our comparison baseline. To isolate the effects of inherent performance changes from those of the dataset, we also train Qwen2.5-7B on the dataset without any prior prompts--thus without format instructions and without a format reward; accuracy reward maxed at 1.0. This setup serves as our No PP baseline to distinguish dataset effects from those of pPE approaches.

We select Qwen2.5-7B, a base model, to follow the R1-Zero \citep{deepseekai2025deepseekr1incentivizingreasoningcapability} approach and mitigate confounding factors from instruction tuning of instruct models; evaluation of the instruct variant is left for future work. Although prompting base models with iPE approaches has become less common in recent years due to the prevalence of instruct models, prior studies introducing the iPE methods \citep{NEURIPS2022_9d560961,wang-etal-2023-plan,chen2023program,liu-etal-2022-generated,taveekitworachai-etal-2024-null} considered here have shown it to be effective with base models.

To assess generalization under budget constraints, we conduct scaled-down experiments along two dimensions: model size and model family. For model size, we train Qwen2.5-3B with \texttt{<think>} and \texttt{<plan>}, as it belongs to the same family as Qwen2.5-7B from the main experiment and allows direct size comparison. For model family, we evaluate Llama 3.1-8B \citep{grattafiori2024llama3herdmodels} with \texttt{<think>} and \texttt{<plan>}, chosen for its comparable size to Qwen2.5-7B, and Qwen2.5-Coder-7B \citep{hui2024qwen25codertechnicalreport} with \texttt{<think>} and \texttt{<code>}, included to examine differences between a code-specialized model and its base counterpart. We prioritize \texttt{<plan>} as the main comparator due to its distinct behaviors during and after RFT, while \texttt{<code>} probes domain-specific specialization.

\subsection{Evaluation}\label{sec:eval}

We evaluate all models and prompting methods via quantitative and qualitative analyses.

\paragraph{Quantitative benchmarks}
Although our training set is math-only, we also evaluate all models on non-mathematical benchmarks to assess generalization. We report average accuracy across the following: \textit{Mathematical reasoning:} AIME2024 (AIME) \citep{li2024numinamath}, AMC12 ’22–’23 (AMC) \citep{li2024numinamath}, and MATH-500 (MATH) \citep{hendrycks2021measuring}; \textit{Coding:} HumanEval+ (HE+) base and extra sets \citep{evalplus}; \textit{Question answering:} GPQA-Diamond (GPQA) \citep{rein2024gpqa}. Additional details are provided in \Cref{sec:quan_eval}.

\paragraph{Qualitative analysis}
We analyze differences across: (1) \textit{Training dynamics}, (2) \textit{Average response length}, (3) \textit{Ratio of four fundamental cognitive behaviors} \citep{gandhi2025cognitivebehaviorsenableselfimproving}, and (4) \textit{Ratio of behavior patterns specific to each of the five pPE categories}. Four fundamental cognitive behaviors are (i) Verification: identifying errors; (ii) Backtracking: proposing an alternative approach; (iii) Subgoal setting: generating intermediate steps; and (iv) Backward chaining: reasoning from the result to inputs. For (3) and (4), we employ the LM-based classification framework of \citet{gandhi2025cognitivebehaviorsenableselfimproving} to automatically classify model responses. Further details are in \Cref{sec:qual_eval}.

\section{Results and Findings}\label{sec:results}

In this section, we present and discuss results from our experiments, as described the setup in \Cref{sec:exp_setup}. Our objective is to answer the core question posed earlier: \emph{whether and how different pPE approaches can guide LMs to internalize distinct behaviors during RFT}. To address this question, we examine three key aspects:
\begin{enumerate}
    \item \textbf{Performance impact:} Do different pPE approaches lead to measurable improvements over the baseline and their iPE counterparts? Do they result in distinct performance gains across tasks, or do they converge to similar outcomes?
    \item \textbf{Behavioral differences:} Do different pPE approaches induce differences in fundamental cognitive behaviors and elicited generation patterns? Do the behavioral profiles of pPE-trained models align with those observed under iPE?
    \item \textbf{Generalization:} How well do pPE approaches generalize across model sizes and families?
\end{enumerate}

The following subsections address each of these aspects in detail. Additional and detailed results, including results from the generalization study, are presented in \Cref{sec:additional_results}.

\subsection{Performance Impact}\label{sec:performance_impact}

\begin{table}[htbp]
  \centering
  \tiny
  \begin{tabular}{lrrrrr|r}
      \toprule
      \textbf{Model} 
        & \textbf{AIME} 
        & \textbf{AMC} 
        & \textbf{GPQA} 
        & \textbf{MATH} 
        & \textbf{HE+} 
        & \textbf{Avg.} \\
      \midrule
        \textbf{Qwen2.5-7B}      
        & 13.33 
        & 37.35 
        & 24.24 
        & 55.60 
        & 72.60 
        & 40.62 \\
      \midrule
      \multicolumn{7}{l}{\textbf{\underline{iPE}}} \\
      \quad Think                 
        & 10.00 
        & 31.33 
        & 24.24 
        & 56.00 
        & \underline{75.00} 
        & 39.31 \\
      \quad Plan                  
        & 10.00 
        & 30.12 
        & 24.24 
        & 51.20 
        & 73.80 
        & 37.87 \\
      \quad Code                  
        & 13.33 
        & 26.51 
        & 24.24 
        & 51.40 
        & 72.00 
        & 37.50 \\
      \quad Knowledge             
        & 20.00 
        & 25.30 
        & 24.24 
        & 59.60 
        & 72.00 
        & 40.23 \\
      \quad Examples              
        & 16.67 
        & 32.53 
        & 24.24 
        & 56.80 
        & 0.00 
        & 26.05 \\
      \midrule
      \multicolumn{7}{l}{\textbf{\underline{RFT}}} \\
      \quad No PP & \textbf{26.67} 
        & 37.35 
        & 21.21
        & 70.40
        & 73.80
        & 45.41 \\
      \midrule
      \multicolumn{7}{l}{\textbf{\underline{pPE}}} \\
      \quad Think   
        & 20.00 
        & 43.37 
        & \underline{28.28} 
        & \textbf{73.20} 
        & 70.10 
        & \underline{46.99} \\
      \quad Plan    
        & 20.00 
        & \underline{44.58} 
        & 24.75 
        & 69.60 
        & 68.90 
        & 45.57 \\
      \quad Code    
        & 16.67 
        & \textbf{46.99} 
        & 25.25 
        & 66.20 
        & \textbf{78.00} 
        & 46.62 \\
      \quad Knowledge
        & 16.67 
        & 37.35 
        & 21.72 
        & 71.00 
        & 73.20 
        & 43.99 \\
      \quad Examples
        & 20.00 
        & 43.37 
        & \textbf{30.81} 
        & \underline{71.20} 
        & 72.60 
        & \textbf{47.60} \\
      \bottomrule
    \end{tabular}
  \caption{Benchmark accuracy (\%) of Qwen2.5-7B when prompted with different iPE or RFT with different pPE approaches across five benchmarks. \textbf{No PP} represents a baseline trained with RFT without any prior prompts. \textbf{Bold} indicates the best performance per column; \underline{underlined} indicates the second best per column.}
  \label{tab:main-results}
\end{table}


\Cref{tab:main-results} presents the performance of Qwen2.5-7B when prompted with different iPE approaches or fine-tuned using RFT with different pPE approaches across benchmarks. Notably, \emph{all iPE approaches result in lower average performance compared to the base model}. For instance, under the null-example utilization approach, iPE fails to generate parsable code during HE+ evaluation, yielding 0.00.

In contrast, \emph{all post-RFT models--regardless of the pPE approach--achieve performance improvements over the base model}. Part of these gains can be attributed to the dataset itself: training without a prior prompt (No PP) raises AIME from 13.33 to 26.67 and MATH from 55.60 to 70.40, outperforming every iPE variant and even all pPE variants on these two math-heavy benchmarks. This indicates that the dataset and RFT alone drive substantial improvements in mathematical reasoning.

The influence of prior prompts becomes more apparent on other benchmarks. On AMC, No PP provides no improvement (37.35 to 37.35), yet every pPE variant surpasses it, with code-based reasoning reaching 46.99. Several pPE approaches also improve GPQA, and the code-based approach raises HE+ from 73.80 to 78.00. Importantly, all pPE variants except knowledge recall outperform No PP in average performance, demonstrating that prior prompts exert an independent effect beyond dataset-driven gains.

The widely used reasoning approach, \texttt{<think>}, serves as a strong pPE baseline and delivers substantial gains (+6.37 points). Surprisingly, the null-example utilization approach, which performs worst under iPE, achieves the highest average improvement (+6.98 points) after RFT--surpassing \texttt{<think>}. Notably, while the null-example iPE approach fails entirely on HE+, its pPE counterpart maintains strong performance on that benchmark. Conversely, the knowledge recall approach, which yields the best iPE performance, produces the weakest results in the pPE setting. These contrasts underscore that \emph{performance trends in iPE do not directly translate to pPE}, highlighting the fundamentally different mechanisms underlying inference-time prompting and RFT.

Finally, as with iPE, pPE methods exhibit diverse benchmark-specific effects. The code-based reasoning approach, while expectedly excelling on HE+, also delivers the strongest AMC score. In contrast, knowledge recall fails to provide meaningful gains on GPQA and even underperforms relative to the base model. Together, these results suggest that \emph{the impact of pPE is more nuanced than simply aligning a domain-specific prompt with a domain-specific task}. We leave further investigation of these dynamics to future work.

\subsection{Behavioral Differences}\label{sec:behavioral_differences}


\begin{figure}[htbp]
  \centering
  \includegraphics[width=\linewidth]{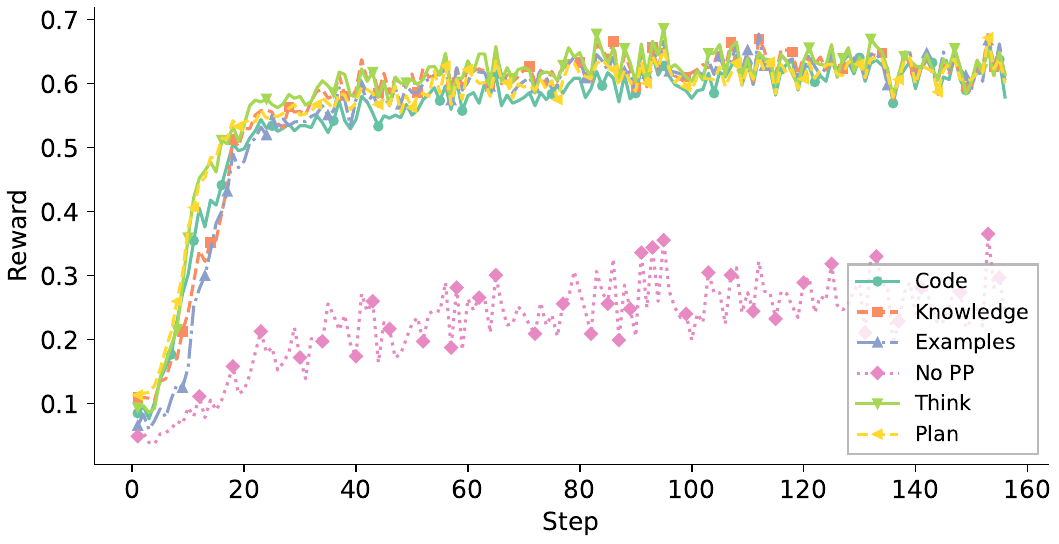}
  \caption{The reward progression of Qwen2.5-7B during RFT exhibits similar trends--an initial climb followed by fluctuations--across all pPE approaches, except for No PP, which yields lower rewards as it focuses only on accuracy without the format component.}
  \label{fig:7b_reward}
\end{figure}

\begin{figure}[htbp]
  \centering
  \includegraphics[width=\linewidth]{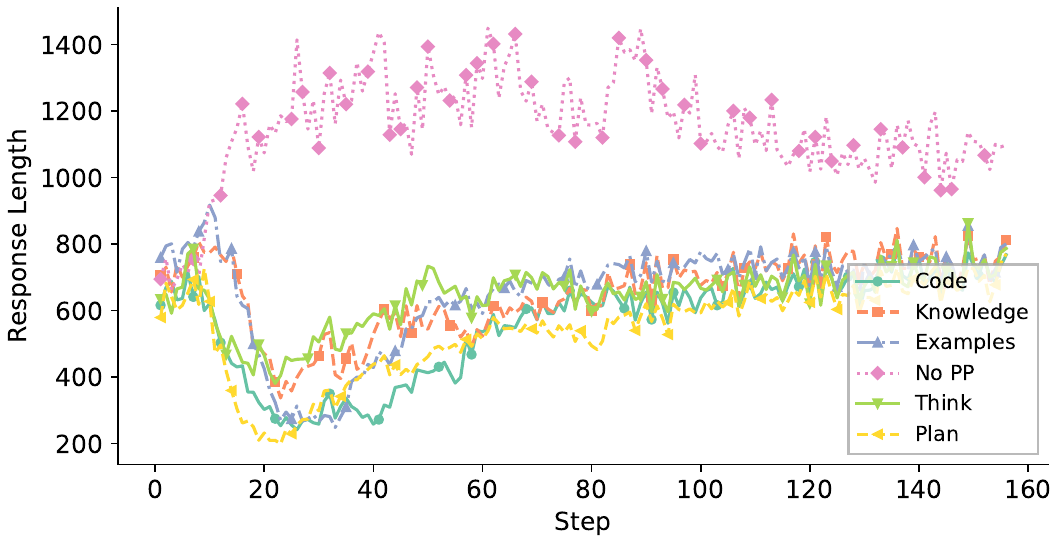}
  \caption{Evolution of the average response length for Qwen2.5-7B during RFT shows an initial drop followed by gradual recovery across pPE approaches, whereas the No PP baseline maintains a higher and more stable response length throughout.}
  \label{fig:7b_response_length}
\end{figure}

\paragraph{Training dynamics} \Cref{fig:7b_reward,fig:7b_response_length} show the reward and average response length dynamics during RFT for each pPE approach, respectively. For all pPE variants, the training curves are highly consistent: reward increases sharply in the first 20 steps, likely reflecting the model learning to follow format constraints, and then enters a steadier phase with minor fluctuations. This phase coincides with a gradual recovery of response length after an initial drop, suggesting that the model begins to exploit a larger token budget in pursuit of higher rewards.

The No PP baseline, however, exhibits markedly different dynamics. In reward progression, it fails to reach the same level as pPE approaches, as it optimizes only for accuracy without benefiting from a format reward. In response length, it shows consistently longer outputs and a relatively steady trend rather than the sharp dip-and-recovery pattern observed in pPE approaches.

These results show that \emph{pPE affects not only final model performance but also the training process itself}. Models trained with pPE share consistent dynamics--rapid reward gains followed by stabilization, along with a dip and recovery in response length--whereas the No PP baseline follows a completely different trajectory, with lower rewards and longer responses. This contrast highlights that the inclusion of a prior prompt fundamentally shapes how the model learns during RFT.

At the same time, other factors such as the training algorithm \citep{yu2025dapoopensourcellmreinforcement,liu2025understandingr1zeroliketrainingcritical}, the base model family \citep{zeng2025simplerlzooinvestigatingtamingzero}, and hyperparameter choices remain important determinants of training behavior. Finally, the divergence between training dynamics and final benchmark results suggests that \emph{metrics like reward progression and response length should not be relied upon as predictors of final performance}.

\begin{table}[htbp]
  \centering
  \tiny
  \begin{tabular}{lrrrrr}
    \toprule
    \textbf{Model}
      & \textbf{AIME}
      & \textbf{AMC}
      & \textbf{GPQA}
      & \textbf{MATH}
      & \textbf{Avg.} \\
    \midrule
    \textbf{Qwen2.5-7B}
      & 1416.80
      & 1352.54
      & 534.29
      & 841.74
      & 1036.34 \\
    \midrule
    \multicolumn{6}{l}{\textbf{\underline{iPE}}} \\
    \quad Think
      & 2512.17
      & 1367.69
      & 534.29
      & 804.85
      & 1304.75 \\
    \quad Plan
      & 1662.57
      &  644.90
      & 534.29
      & 540.98
      &  845.69 \\
    \quad Code
      &  641.07
      &  953.51
      & 534.29
      & 635.09
      &  690.99 \\
    \quad Knowledge
      & 1406.30
      & 1237.22
      & 534.29
      & 780.31
      &  989.53 \\
    \quad Examples
      & 2274.17
      & 1316.12
      & 534.29
      & 752.60
      & 1219.30 \\
    \midrule
    \multicolumn{6}{l}{\textbf{\underline{RFT}}} \\
      \quad No PP & 2902.97 
        & 1543.40
        & 982.79
        & 850.33
        & 1569.87 \\
      \midrule
    \multicolumn{6}{l}{\textbf{\underline{pPE}}} \\
    \quad Think
      & 2042.70
      & 1024.96
      & 476.86
      & 612.10
      & 1039.16 \\
    \quad Plan
      & 1685.17
      & 1085.47
      & 476.47
      & 601.18
      &  962.07 \\
    \quad Code
      & 1657.47
      &  836.28
      & 492.44
      & 690.42
      &  919.15 \\
    \quad Knowledge
      & 2015.57
      & 1082.96
      & 587.10
      & 626.45
      & 1078.02 \\
    \quad Examples
      & 1136.20
      &  831.98
      & 442.48
      & 685.79
      &  774.11 \\
    \bottomrule
  \end{tabular}
  \caption{Average response length, i.e., number of tokens, of Qwen2.5-7B when prompted with different iPE or RFT with different pPE approaches.}
  \label{tab:main-response-length}
\end{table}

\paragraph{Average response length} \Cref{tab:main-response-length} reports the average response length, measured as the mean number of generated tokens during quantitative evaluation. We find that reasoning and null-example utilization iPE approaches already elicit longer responses compared to the base model. After RFT, average response length generally increases further, though the No PP baseline produces by far the longest responses across all benchmarks, despite not achieving the strongest performance. In contrast, pPE variants yield shorter and more varied response lengths.

Interestingly, the null-example utilization pPE approach achieves the highest overall performance while producing some of the shortest responses on average, making it the most efficient in terms of test-time compute. By comparison, the reasoning pPE approach also reduces response length relative to its iPE counterpart while still delivering strong performance gains.

These results indicate that \emph{different pPE approaches shape not only the performance but also the efficiency of post-RFT models}. In particular, these results highlight \emph{pPE as a practical tool for influencing the trade-off between model accuracy and computational efficiency}.

\begin{figure*}[htbp]
\begin{center}
\includegraphics[width=\linewidth]{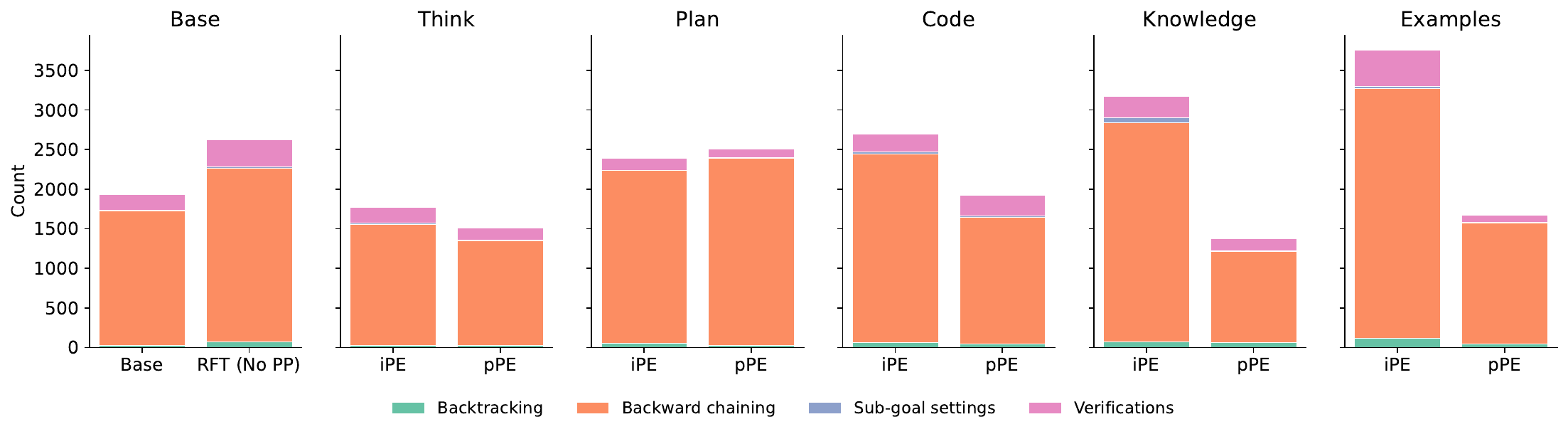}
\end{center}
\caption{Ratio of the four fundamental cognitive behaviors--backtracking, backward chaining, subgoal setting, and verification--across different prompting (iPE) and RFT (pPE) approaches with Qwen2.5-7B. Backward chaining dominates across setups, especially under iPE.}
\label{fig:main_four_behav}
\end{figure*}

\paragraph{Four fundamental behaviors} \Cref{fig:main_four_behav} shows the ratio of four fundamental cognitive behaviors in responses from the quantitative evaluation, both when prompted with iPE approaches and after RFT with pPE approaches. We observe that \emph{backward chaining is the most prominent behavior across all models}--regardless of whether iPE or pPE is used--and is already present in the base model. Interestingly, the base model with RFT without any prior prompts displays even higher levels of backward chaining, backtracking, and verifications than the raw base, indicating that RFT with the math-only dataset alone encourages more cognitive behaviors.

In general, iPE approaches increase the frequency of backward chaining, while pPE approaches tend to reduce it, with the exception of the planning approach. More broadly, pPE approaches tend to decrease the overall presence of all four fundamental cognitive behaviors compared to iPE. Importantly, the ratio of these behaviors does not correlate well with final model performance. However, these ratios remain useful for highlighting how fundamental cognitive behaviors shift post-RFT, and for differentiating between pPE approaches based on their behavioral profiles.

We speculate that this behavior classification framework--originally developed to analyze reasoning models \citep{gandhi2025cognitivebehaviorsenableselfimproving}--may not generalize well to models trained with different pPE paradigms, which may incentivize different forms of fundamental behavior beyond those captured by the current classification framework.

\begin{figure}[htbp]
\begin{center}
\includegraphics[width=\linewidth]{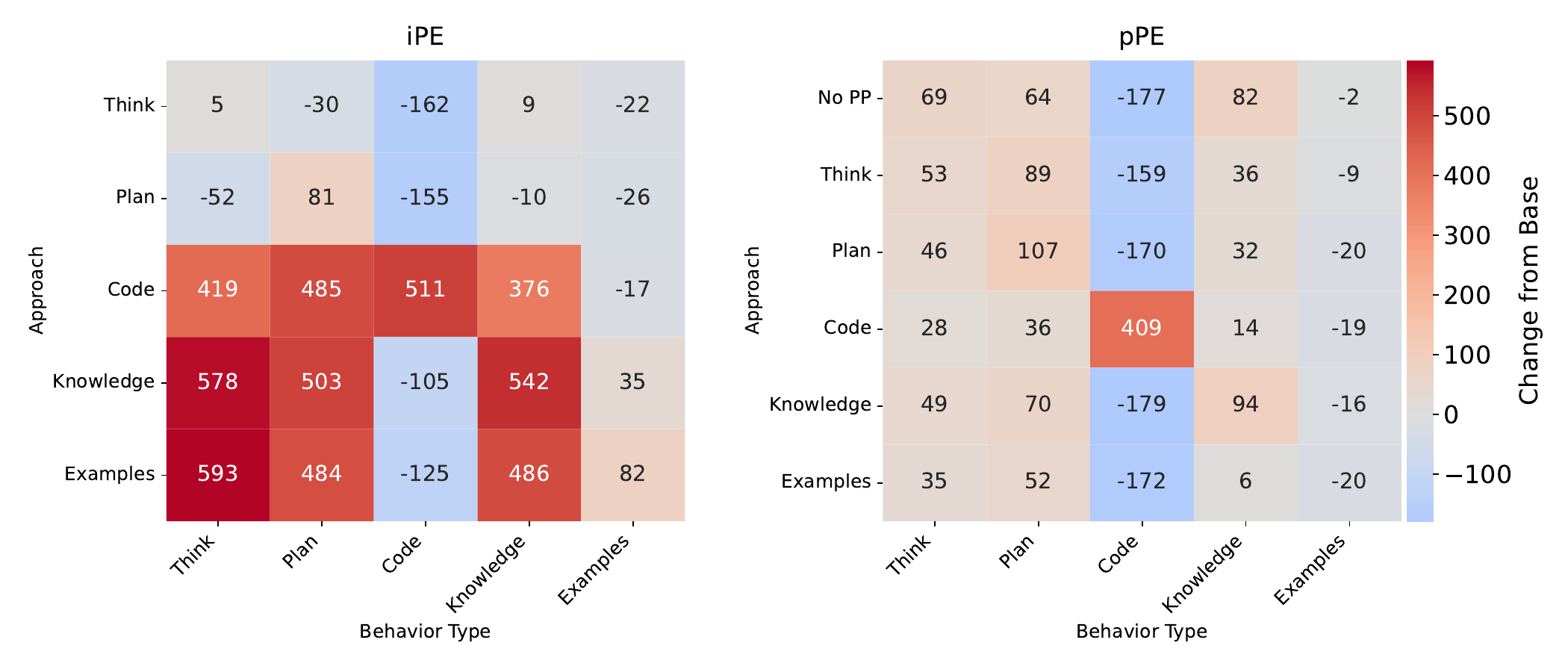}
\end{center}
\caption{Ratio of five elicited behavior categories--reasoning, planning, code-based reasoning, knowledge recall, and null-example utilization--as observed when behaviors are elicited through prompting with different iPE approaches (left) and after RFT with different pPE approaches (right).}
\label{fig:main_five_template}
\end{figure}

\paragraph{Five elicited behaviors} \Cref{fig:main_five_template} shows changes in the frequency of five elicited behaviors when models are prompted or trained using iPE or pPE approaches, relative to the base model under zero-shot prompting. We observe that most iPE approaches--with the exception of reasoning and planning--elicit high levels of reasoning, planning, and knowledge recall behaviors. In contrast, post-RFT behavior patterns are more targeted: \emph{post-RFT models tend to show their largest gains in the behavior aligned with the specific pPE approach they were trained on, with the notable exception of the null-example utilization approach.} For instance, the planning pPE yields the strongest increases in planning, while the code pPE uniquely boosts code-related behaviors--consistent with expectations.

The No PP baseline shows that RFT alone increases reasoning, planning, and knowledge recall behaviors relative to the zero-shot baseline. This indicates that RFT with math-only datasets already incentivizes models to exhibit these behaviors.

Finally, each pPE still induces a distinct distribution across the five behaviors. Notably, the null-example utilization pPE yields the fewest knowledge recall instances, yet achieves the highest performance gains on GPQA. Furthermore, it also exhibits the lowest number of null-example behavior instances--in contrast to its iPE counterpart and to our expectations. This suggests that \emph{a pPE approach may not always result in the model exhibiting the anticipated behavior}. Instead, the model may discover more effective behavior patterns during RFT, independent of the specific pPE approach.

\paragraph{Qualitative behaviors} We present qualitative examples of generated responses in \Cref{sec:qual_examples}. We observe that \emph{post-RFT models are generally able to produce behaviors aligned with the pPE approaches}. For example, the planning pPE approach results in models that generate a numbered list of steps to solve the problem and then execute them. The reasoning pPE approach leads to step-by-step reasoning, while the knowledge recall pPE approach elicits definitions and formulas relevant to solving the task. Interestingly, the null-example utilization pPE approach somewhat resembles the behavior of the reasoning pPE model, despite its differences in performance trends.

We also observe that Qwen2.5-7B tends to prefer natural language reasoning over code-based reasoning during RFT. Specifically, under the code-based reasoning pPE approach, the model frequently generates natural language reasoning, followed by a statement such as:

\begin{quote}
\texttt{<code>}\\
\# We don't need to write any Python code since the problem is solved analytically.\\
\texttt{</code>}
\end{quote}

This stands in contrast to the code-specialized model, which, as shown in the qualitative examples, relies more heavily on code generation as part of its problem-solving process. \emph{These behavioral differences among post-RFT models suggest that RFT with different pPE approaches can be used to steer models toward exhibiting distinct, desired behaviors}--similar to RLHF \citep{NEURIPS2022_b1efde53}. For instance, it is possible to train a plan-generating model by applying RFT with a prior prompt that elicits plan generation. However, for such a model to be effective, the plan must not only be valid but also executable in a way that achieves a high reward. In this context, the reward signal serves as a proxy for plan quality. We note that all qualitative analyses presented here are preliminary and focus on observed differences in behaviors. A deeper analysis of the mechanisms by which pPE or RFT influence the trained model's behaviors is beyond the scope of this work. We further discuss the implications, extensions, and applications of pPE for RFT in \Cref{sec:discussions}, as well as limitations and possible extensions in \nameref{sec:limitations}.

\subsection{Generalization}\label{sec:generalization}

\begin{table}[htbp]
	\centering
	\tiny
    \setlength\doublerulesep{0.5pt}
    \setlength{\tabcolsep}{5pt}
	\begin{tabular}{lrrrrr|r}
		\toprule
		\textbf{Model} 
		& \textbf{AIME} 
		& \textbf{AMC} 
		& \textbf{GPQA} 
		& \textbf{MATH} 
		& \textbf{HE+} 
		& \textbf{Avg.} \\
		\midrule
		\textbf{Qwen2.5 3B}                    
		& \textbf{13.33} 
		& \textbf{24.10} 
		& \underline{9.60} 
		& \textbf{49.40} 
		& \textbf{62.80} 
		& \textbf{31.85} \\
		\midrule
		\multicolumn{7}{l}{\textbf{\underline{iPE}}} \\
		\quad Think                 
		& \textbf{13.33}
		& \underline{22.89}
		& \underline{9.60}
		& \underline{37.20}
		& \underline{61.00}
		& \underline{28.80} \\
		\quad Plan                  
		& \textbf{13.33} 
		& 15.66 
		& \underline{9.60} 
		& 35.20 
		& 60.40 
		& 26.84 \\
		\midrule
		\multicolumn{7}{l}{\textbf{\underline{pPE}}} \\
		\quad Think    
		& 10.00
		& 12.05
		& \textbf{11.11}
		& 28.40
		& \underline{61.00}
		& 24.51 \\
		\quad Plan     
		&  0.00 
		&  0.00 
		&  7.07 
		&  0.00 
		& 59.10 
		& 13.23 \\
		\midrule\midrule
		\textbf{Llama 3.1-8B}                    
		  & 0.00             & 1.20             & 0.00 & 5.00             & 31.70             & 7.58              \\
		\midrule
		\multicolumn{7}{l}{\textbf{\underline{iPE}}} \\
		\quad Think                 
		  & \underline{3.33} & 3.61             & 0.00 & 8.00             & 31.70             & 9.33              \\
		\quad Plan                  
		  & \textbf{6.67}    & \underline{4.82} & 0.00 & 6.80             & 31.70             & \underline{10.00} \\
		\midrule
		\multicolumn{7}{l}{\textbf{\underline{pPE}}} \\
		\quad Think   
		  & \underline{3.33} & \textbf{6.02}    & 0.00 & \textbf{9.60}    & \underline{32.30} & \textbf{10.25}    \\
		\quad Plan    
		  & \underline{3.33} & 2.41             & 0.00 & \underline{8.40} & \textbf{32.90}    & 9.41              \\
		\midrule\midrule
		\textbf{Qwen2.5-Coder-7B}                    
		& \underline{6.67} 
		& 15.66 
		& 25.25
		& \underline{27.60} 
		& \textbf{81.10} 
		& 31.26 \\
		\midrule
		\multicolumn{7}{l}{\textbf{\underline{iPE}}} \\
		\quad Think                 
		&  0.00 
		& 10.84 
		& 25.25 
		& 22.20 
		& 75.60 
		& 26.78 \\
		\quad Code                  
		&  3.33 
		& 10.84 
		& 25.25 
		& 12.00 
		& 78.00 
		& 25.88 \\
		\midrule
		\multicolumn{7}{l}{\textbf{\underline{pPE}}} \\
		\quad Think    
		& \textbf{13.33} 
		& \textbf{37.35} 
		& \underline{26.26} 
		& \textbf{68.80} 
		& \underline{78.70} 
		& \textbf{44.89} \\
		\quad Code     
		& \textbf{13.33} 
		& \underline{18.07} 
		& \textbf{34.85} 
		& 26.80 
		& 75.00 
		& \underline{33.61} \\
		\bottomrule
	\end{tabular}
	\caption{Benchmark accuracy (\%) of Qwen2.5-3B, Llama 3.1-8B, and Qwen2.5-Coder-7B when prompted with different iPE or RFT with different pPE approaches across five benchmarks. \textbf{Bold} indicates the best performance per column under the same base model; \underline{underlined} indicates the second best per column under the same base model.}
	\label{tab:generalization-results}
\end{table}


\paragraph{Performance impact} \Cref{tab:generalization-results} shows the performance of Qwen2.5-3B, LLaMA 3.1-8B, and Qwen2.5-Coder-7B, which serve as representative models for our generalization studies. Additional accompanying results--including training dynamics and behavior classification--are available in \Cref{sec:additional_results}. We observe that the reasoning pPE approach, i.e., \texttt{<think>}, is consistently robust across model families and sizes. This is likely due to its alignment with behavioral patterns already familiar to models from prior fine-tuning on CoT-like data \citep{JMLR:v25:23-0870}. In contrast, \emph{smaller or weaker model families show limited success with non-reasoning pPE approaches}, aligning with findings from \citep{zeng2025simplerlzooinvestigatingtamingzero} that such models benefit less from reasoning RFT.

\paragraph{Behavioral differences} We also observe instances of reward hacking when RFT is applied with the planning pPE approach in Qwen2.5-3B and LLaMA 3.1-8B. In these cases, the models output only correctly formatted responses in order to maximize the format reward, while neglecting further exploration of the accuracy reward (see \Cref{fig:3b_reward,fig:3b_response_length,fig:llama31_reward,fig:llama31_response_length}). Another notable observation is that the code-specialized Qwen2.5-Coder-7B model is more effective at exhibiting code-based reasoning behaviors--under both iPE and pPE--compared to Qwen2.5-7B. This illustrates how different model families can influence the behaviors exhibited after RFT \citep{zeng2025simplerlzooinvestigatingtamingzero}. Nevertheless, both reasoning and code-based reasoning pPE approaches improve performance over the baseline, although the reasoning pPE approach achieves the highest performance. Still, pPE demonstrates measurable success over iPE in steering model behavior post-RFT, as illustrated in \Cref{fig:3b_five_template,fig:llama_five_template,fig:coder_five_template}. These findings suggest that \emph{pPE generalizes reliably in stronger, behaviorally aligned models, while less capable models are more prone to reward hacking or fail to internalize the intended behaviors}.

\section{Conclusions}\label{sec:conclusions}

This paper investigates the impact of pPE in the context of RFT by evaluating five pPE approaches inspired by iPE: reasoning, planning, code-based reasoning, knowledge recall, and null-example utilization. While these approaches often degrade performance when applied only at inference time (iPE), incorporating them during RFT (pPE) consistently improves performance relative to the base model. In particular, null-example utilization proves more effective than the reasoning approach for enhancing downstream task performance.

Beyond these performance impact, different pPE approaches induce distinct behavioral patterns in the fine-tuned models. For example, models trained with the planning pPE approach tend to exhibit a ``plan-and-solve'' behavior, i.e., generating a list of steps before execution. Finally, we explore the generalization of pPE approaches across model sizes and families. We hope this study will inspire further research into the role of pPE in RFT, especially given the extensive literature on iPE.

\section*{Limitations}\label{sec:limitations}

Due to computational resource constraints, we were unable to conduct experiments with larger model sizes, larger datasets, or a higher number of steps. As a result, the behavioral trends observed in this study remain inconclusive for larger models where emergent abilities \citep{wei2022emergent}--known to appear only at scale--may lead to different outcomes. While we believe many of our findings will generalize across model sizes (as is often the case in iPE studies), this assumption remains to be validated. In contrast, smaller models may not capture the complexity or expressiveness of their larger counterparts due to their lower capacity and the limited potential of pPE, similar to iPE.

Additionally, we fixed the training data domain (mathematics), the reinforcement learning algorithm (GRPO), and other experimental configurations to isolate the effect of pPE approaches. Future work should investigate how different domains, RL algorithms, and reward schemes interact with pPE. We conjecture that, as with iPE, once a model demonstrates the ability to exhibit structured behaviors, those behaviors will generalize across architectures and settings. However, further studies are necessary to confirm this generalization in a broader context of RFT.

\section*{Ethical Considerations}

Prompting language models to elicit specific behaviors is inherently unpredictable due to their stochastic nature \citep{NIPS2000_728f206c}. RFT, which aims to amplify specific generation patterns for improved performance, may also unintentionally reinforce undesirable or unsafe behaviors--especially those that were already latent in the pretrained model.

We strongly recommend integrating established alignment techniques \citep{grattafiori2024llama3herdmodels,bai2022constitutionalaiharmlessnessai,dai2024safe} and safety measures \citep{zeng2025shieldgemma2robusttractable,inan2023llamaguardllmbasedinputoutput}, as prior studies \citep{deepseekai2025deepseekr1incentivizingreasoningcapability,seed2025seedthinkingv15advancingsuperbreasoning} have shown that such safeguards remain effective even after RFT. As with iPE, models in dynamic or open-ended environments are vulnerable to misuse. For example, they may be exposed to malicious prompts \citep{liu2024promptinjectionattackllmintegrated} or poisoned data \citep{zhao2025datapoisoningdeeplearning} during RFT, leading to unexpected or concerning behaviors.

Furthermore, the pPE framework proposed in this study can be extended to alignment and safety-focused training, similar to recent efforts in \textit{deliberative alignment} \citep{guan2025deliberativealignmentreasoningenables}. To mitigate risks, we recommend safeguards such as prompt auditing, robust reward design, safe rollout filtering, and post-training alignment steps--especially when applying RFT in safety-critical or user-facing applications.

\bibliography{custom,anthology}

\appendix

\section{Related Work}

\subsection{Reinforcement Fine-Tuning (RFT)}

Reinforcement learning (RL) has become a common post-training method for large language models (LLMs) \citep{kumar2025llmposttrainingdeepdive}. One prominent RL approach is reinforcement learning from human feedback (RLHF) \citep{NEURIPS2022_b1efde53}, where a reward model--trained on human preference comparisons--predicts scalar scores for model outputs. This enables optimization of model behavior toward human-aligned responses, typically using Proximal Policy Optimization (PPO) \citep{schulman2017proximalpolicyoptimizationalgorithms}.

A recent shift in RL for LLM post-training is the introduction of \textit{reinforcement learning with verifiable rewards} (RLVR), also known as \textit{reinforcement fine-tuning} (RFT). First introduced by \citet{olmo20252olmo2furious}, RFT replaces the reward model with task-specific, rule-based reward functions for domains with verifiable answers such as mathematics, logic, and code. This not only improves performance but also eliminates the need to train a separate reward model and maintain it during training.

RFT gained widespread attention following the release of DeepSeek-R1-Zero \citep{deepseekai2025deepseekr1incentivizingreasoningcapability}, which extends the RLVR paradigm by incorporating two key modifications: (1) replacing PPO with Group Relative Policy Optimization (GRPO) \citep{shao2024deepseekmathpushinglimitsmathematical} to eliminate the need for a separate value model, reducing compute cost; and (2) introducing a \textbf{prior prompt} to elicit reasoning behavior during training. While the former has received significant attention, the latter--prior prompt--remains largely understudied.

The core components of RFT include: (1) the RL algorithm, (2) base language model, (3) training dataset, (4) reward function, and (5) prior prompt. Recent studies have explored improvements in RL algorithms (e.g., Dr. GRPO \citep{liu2025understandingr1zeroliketrainingcritical}, DAPO \citep{yu2025dapoopensourcellmreinforcement}, VAPO \citep{yue2025vapoefficientreliablereinforcement}), base model effects \citep{zeng2025simplerlzooinvestigatingtamingzero,gandhi2025cognitivebehaviorsenableselfimproving}, and expanding verifiable tasks such as logic \citep{xie2025logicrlunleashingllmreasoning}, coding \citep{wei2025swerladvancingllmreasoning}, and function calling \citep{feng2025retoolreinforcementlearningstrategic}. However, the \textit{role} of prior prompts has received little attention. Aside from one study noting their effect on training stability \citep{xie2025logicrlunleashingllmreasoning,hu2025openreasonerzeroopensourceapproach}, prompt design in RFT remains significantly underexplored. Given the importance of prompting in inference-time settings (iPE), we argue that pPE deserves focused study as a core axis of RFT.

\subsection{Inference-Time Prompt Engineering (iPE)}

iPE has seen rapid development since the introduction of ChatGPT \citep{sahoo2025systematicsurveypromptengineering}. iPE refers to techniques for prompting LLMs to produce desirable outcomes. A prominent direction in iPE is reasoning-centric prompting, with the seminal work being chain-of-thought (CoT) prompting \citep{NEURIPS2022_9d560961}, which uses in-context examples to demonstrate multi-step reasoning. This was later extended by zero-shot CoT \citep{kojima2022large} prompting, where a simple phrase like ``Let's think step by step.'' is sufficient to elicit similar behavior from capable LLMs.

Since then, many variants have emerged to elicit a range of intermediate reasoning patterns--beyond just step-by-step reasoning. These include prompting for planning \citep{wang-etal-2023-plan}, code generation \citep{chen2023program}, knowledge recall \citep{liu-etal-2022-generated}, and hallucination induction \citep{taveekitworachai-etal-2024-null}.

We note a conceptual parallel between iPE and pPE: both \textit{initially} focused on reasoning, but iPE has since broadened to include diverse useful behaviors \citep{sahoo2025systematicsurveypromptengineering}. Motivated by this, our study extends RFT by incorporating a range of iPE-inspired prompting strategies as prior prompts. We aim to investigate whether these paradigms, when moved from inference to training time, yield corresponding behavioral changes during RFT.

\section{Additional Discussions}\label{sec:discussions}

\subsection{Domain generalization}\label{sec:domain_generalization}

We observe that, although only mathematical problems is used during training, performance improvements often extend to other domains--as seen in GPQA and HE+ in \Cref{tab:main-results,tab:generalization-results}. This demonstrates the robustness of the RFT approach in general and suggests that RFT may function more as a mechanism for discovering useful generation patterns than for infusing the model with new knowledge. We expect broader performance generalization to emerge with more diverse training data, such as by incorporating code or logic problems.

\subsection{pPE for RFT}\label{sec:ppe_for_rft}

As demonstrated in this study, the importance of pPE in RFT is analogous to the role of iPE for LMs. Prompts play a critical role in conditioning the base model's generation, which in turn affects the trajectories sampled during RFT--ultimately leading to distinct post-training behaviors.

This also implies that properties known to affect LMs during inference, such as sensitivity to prompt wording \citep{kojima2022large, shanahan2023}, formatting \citep{sclar2024quantifying, tang-etal-2024-struc}, and prompt order \citep{taveekitworachai-etal-2024-null, min-etal-2022-rethinking}, can similarly influence RFT outcomes. However, our results in \Cref{sec:performance_impact} suggest that insights from iPE do not directly translate to RFT--reinforcing the need for targeted study of pPE. That said, the model's instruction-following capabilities can still be leveraged to incentivize distinct behavioral patterns through carefully designed prior prompts.

\subsection{Beyond Reasoning Models}\label{sec:beyond_reaosning_model}

We discuss here several promising directions enabled by pPE, inspired by advances in iPE:

\paragraph{Not only think, plan, code, and recall knowledge}
The reasoning trace itself can be an important vehicle for interpretability and user trust \citep{NEURIPS2022_9d560961}. Given that we can elicit distinct reasoning styles, we may tailor them to user preferences or application requirements. For instance, an LLM could reason in a self-talk style using \texttt{<dialogues>} tags, or imitate a specific style using few-shot demonstrations \citep{NEURIPS2020_1457c0d6}. 

Recent work on chain-of-draft prompting \citep{xu2025chaindraftthinkingfaster} shows that natural language constraints can guide the model to produce shorter but still effective reasoning traces. Such behavior can likely be transferred into pPE settings, especially with models that possess strong instruction-following capabilities. The breadth of iPE research suggests many additional styles--beyond those we studied here--could be explored and reinforced via pPE.

\paragraph{Dynamic pPE}
As shown in \Cref{tab:main-results}, different pPE approaches excel in different domains. Dynamically selecting the prior prompt based on the task or question difficulty could further enhance performance. This idea aligns with the test-time scaling paradigm \citep{zhang2025surveytesttimescalinglarge}, which advocates allocating more resources to harder inputs. 

Furthermore, prior prompts could become part of the RL optimization process, akin to automatic prompt search \citep{ramnath2025systematicsurveyautomaticprompt}. While such approaches increase system complexity, they offer a path toward more adaptive and robust behaviors.

\paragraph{Structured thinking}
Instead of using a single behavior tag (e.g., \texttt{<think>}), we may extend to multi-tag structures (e.g., combining \texttt{<plan>} and \texttt{<code>}) to guide the model through more structured multi-phase reasoning processes. This may be especially beneficial in tasks requiring distinct reasoning modes at different stages (e.g., planning followed by execution).

\paragraph{Incentivizing behaviors through verifiable rewards}
Consider a model trained with the \texttt{<plan>} prompt. During training, it learns to produce a useful plan inside the \texttt{<plan>} tag before solving the problem in \texttt{<answer>}. Because final task accuracy is used as a reward, this implicitly incentivizes the model to generate effective intermediate content. Thus, verifiable rewards can act as a \textit{surrogate signal} for training behaviors--such as a planning or coding--without \textit{direct} supervised signals.

This logic extends to other prompts: if we stop generation after \texttt{</plan>} or \texttt{</code>}, we can re-purpose these models to act as plan generators or code synthesizers. This strategy opens up a broader class of behavioral specialization, where useful intermediate behaviors can be extracted and repurposed for downstream applications—all trained indirectly via RFT.


\section{Prior Prompt Examples}\label{sec:prior_prompt_examples}

In this section, we provide additional two examples of prior prompts used in existing RFT studies to elicit reasoning behavior during RFT. These are \Cref{fig:prior_logicrl,fig:prior_openreasoner}.

\begin{figure}[htbp]
\footnotesize
\centering
\begin{tcolorbox}[title=Logic-RL Prior Prompt, colback=gray!5, colframe=amethyst!75!black]
\input{prompts/prior_examples/logic_rl.txt}
\end{tcolorbox}
\caption{The prompt used during RFT by Logic-RL \citep{xie2025logicrlunleashingllmreasoning}.}
\label{fig:prior_logicrl}
\end{figure}

\begin{figure}[htbp]
\footnotesize
\centering
\begin{tcolorbox}[title=Open-Reasoner-Zero Prior Prompt, colback=gray!5, colframe=amethyst!75!black]
\input{prompts/prior_examples/open_reasoner_zero.txt}
\end{tcolorbox}
\caption{The prompt used during RFT by Open-Reasoner-Zero \citep{hu2025openreasonerzeroopensourceapproach}.}
\label{fig:prior_openreasoner}
\end{figure}

\section{Additional Experimental Setup Details}\label{sec:training_setup}

This section provides additional implementation details of our experimental setup, including prior prompt templates in \Cref{sec:prior_prompts}, training scripts in \Cref{sec:training_script}, reward function design in \Cref{sec:reward_design}, and evaluation details in \Cref{sec:a_eval}.

\subsection{Prior Prompts}\label{sec:prior_prompts}

This section presents the full set of prior prompts used in our experiments, as described in \Cref{sec:exp_setup}. Each prompt was designed to elicit different behavioral styles from the model. These prompts are: \texttt{<think>} (\Cref{fig:prior_think}, for step-by-step reasoning), \texttt{<plan>} (\Cref{fig:prior_plan}, for planning), \texttt{<code>} (\Cref{fig:prior_code}, for reasoning through code), \texttt{<knowledge>} (\Cref{fig:prior_knowledge}, for recalling relevant facts), and \texttt{<examples>} (\Cref{fig:prior_examples}, for utilizing null-examples).

\begin{figure}[htbp]
\footnotesize
\centering
\begin{tcolorbox}[title=Think Prompt, colback=gray!5, colframe=amethyst!75!black]
\input{prompts/prior/think.txt}
\end{tcolorbox}
\caption{The \texttt{<think>} prior prompt, inspired by chain-of-thought (CoT) prompting \citep{NEURIPS2022_9d560961}, encourages the model to reason step by step before concluding with an answer.}
\label{fig:prior_think}
\end{figure}

\begin{figure}[htbp]
\footnotesize
\centering
\begin{tcolorbox}[title=Plan Prompt, colback=gray!5, colframe=amethyst!75!black]
\input{prompts/prior/plan.txt}
\end{tcolorbox}
\caption{The \texttt{<plan>} prior prompt, based on plan-and-solve prompting \citep{wang-etal-2023-plan}, asks the model to explicitly lay out a plan before solving the problem.}
\label{fig:prior_plan}
\end{figure}

\begin{figure}[htbp]
\footnotesize
\centering
\begin{tcolorbox}[title=Code Prompt, colback=gray!5, colframe=amethyst!75!black]
\input{prompts/prior/code.txt}
\end{tcolorbox}
\caption{The \texttt{<code>} prior prompt encourages the model to reason through code, inspired by program-of-thought (PoT) prompting \citep{chen2023program}.}
\label{fig:prior_code}
\end{figure}

\begin{figure}[htbp]
\footnotesize
\centering
\begin{tcolorbox}[title=Knowledge Prompt, colback=gray!5, colframe=amethyst!75!black]
\input{prompts/prior/knowledge.txt}
\end{tcolorbox}
\caption{The \texttt{<knowledge>} prior prompt elicits factual recall relevant to the problem before beginning reasoning, inspired by generated knowledge prompting \citep{liu-etal-2022-generated}.}
\label{fig:prior_knowledge}
\end{figure}

\begin{figure}[htbp]
\footnotesize
\centering
\begin{tcolorbox}[title=Examples Prompt, colback=gray!5, colframe=amethyst!75!black]
\input{prompts/prior/examples.txt}
\end{tcolorbox}
\caption{The \texttt{<examples>} prior prompt draws on null-shot prompting \citep{taveekitworachai-etal-2024-null} to encourage the model to provide illustrative examples before answering.}
\label{fig:prior_examples}
\end{figure}

\subsection{Reward Design}\label{sec:reward_design}

We design our reward function with two equally weighted components: (1) an \textbf{accuracy reward} and (2) a \textbf{format reward}. This setup follows the approach introduced by \citet{xie2025logicrlunleashingllmreasoning}. While some studies suggest the format reward may not be necessary \citep{zeng2025simplerlzooinvestigatingtamingzero}, we find that it is crucial in our setting to ensure the model outputs are well-structured.

\begin{itemize}
    \item \textbf{Accuracy}: The accuracy reward is based on the model's predicted answer, extracted from the content enclosed in \texttt{\textbackslash boxed\{\}}. We use the \texttt{math-verify}\footnote{\url{https://github.com/huggingface/Math-Verify}} package (Apache License 2.0) to check for mathematical equivalence with the ground-truth answer. If the answer is equivalent, the model receives a reward of $0.5$; otherwise, it receives $0.0$.

    \item \textbf{Format}: We adopt a relaxed version of the format reward from the \texttt{open-r1}\footnote{\url{https://github.com/huggingface/open-r1}} (Apache License 2.0) implementation. The reward is given if the response includes \textit{exactly one pair} of the expected XML tags (e.g., \texttt{<think>...</think>} followed by \texttt{<answer>...</answer>}), and the content satisfies basic XML structure constraints, even if the tags are not the only elements in the string. This constraint discourages generation of multiple or malformed tag pairs. If the response satisfies these constraints, a reward of $0.5$ is granted; otherwise, it receives $0.0$.
\end{itemize}

The total reward is the sum of these two components, yielding a final reward in the range $[0, 1]$. This balanced reward design helps incentivize both correct and well-structured responses during RFT.








\subsection{Training Setup and Hyperparameters}\label{sec:training_script}

We use the training script illustrated in \Cref{fig:training_script} for training the models as described in \Cref{sec:training}. All training runs use a single node equipped with 8xH100 GPUs. Across all experiments presented in this paper, we utilized a total of 78 GPU-hours.

We note that both OpenRLHF and vLLM are available under the Apache License 2.0. The Qwen2.5-7B model used in our main experiments is distributed under the Apache License 2.0. For our generalization studies, the Qwen2.5-3B model is distributed under the Qwen Research License, the Llama 3.1-8B model under the Llama 3.1 Community License Agreement, and the Qwen2.5-Coder-7B model under the Apache License 2.0. All of these licenses permit use for research purposes.

\begin{figure}[htbp]
\footnotesize
\centering
\begin{tcolorbox}[title=OpenRLHF Training Script, colback=gray!5, colframe=amethyst!75!black]
\begin{lstlisting}[language=bash]
python3 -m openrlhf.cli.train_ppo_ray \
    --ref_num_nodes 1 \
    --ref_num_gpus_per_node 8 \
    --actor_num_nodes 1 \
    --actor_num_gpus_per_node 8 \
    --vllm_num_engines 8 \
    --vllm_tensor_parallel_size 1 \
    --colocate_all_models \
    --vllm_enable_sleep \
    --vllm_gpu_memory_utilization 0.5 \
    --pretrain "Qwen/Qwen2.5-7B" \
    --remote_rm_url "reward_function.py" \
    --micro_train_batch_size 1 \
    --train_batch_size 64 \
    --micro_rollout_batch_size 8 \
    --rollout_batch_size 64 \
    --n_samples_per_prompt 8 \
    --enable_prefix_caching \
    --max_epochs 1 \
    --prompt_max_len 1024 \
    --max_samples 10000 \
    --generate_max_len 4096 \
    --init_kl_coef 1e-6 \
    --gamma 1.0 \
    --use_kl_loss \
    --kl_estimator k3 \
    --advantage_estimator group_norm \
    --zero_stage 2 \
    --bf16 \
    --actor_learning_rate 5e-7 \
    --prompt_data "user/stillv3" \
    --prompt_split "train" \
    --input_key "query" \
    --label_key "answer" \
    --apply_chat_template \
    --normalize_reward \
    --adam_offload \
    --gradient_checkpointing \
    --flash_attn \
    --packing_samples
\end{lstlisting}
\end{tcolorbox}
\caption{Training script using the OpenRLHF for RFT. This script specifies the model, dataset, GRPO algorithm, reward configuration, and other relevant hyperparameters.}
\label{fig:training_script}
\end{figure}

\subsection{Evaluation}\label{sec:a_eval}

In this section, we provide additional details on quantitative and qualitative evaluation, mentioned in \Cref{sec:eval}.

\subsubsection{Quantitative Analysis}\label{sec:quan_eval}

We evaluate the performance of each trained model using both in- and out-of-domain benchmarks. During evaluation, we consistently prepend the same prior prompt used during training to elicit the trained behaviors. We evaluate once with fixed random seed using pass@1 accuracy. The benchmarks are:

\begin{itemize}
    \item \textbf{Mathematical reasoning}: AIME24 \citep{li2024numinamath}, AMC12 '22–'23 \citep{li2024numinamath}, and MATH-500 \citep{hendrycks2021measuring} are benchmarks used for evaluating mathematical reasoning and serve as our primary in-domain evaluations.
    \item \textbf{Coding}: HumanEval+ (base and extra) \citep{evalplus} is used to evaluate general coding ability and serves as an out-of-domain probe.
    \item \textbf{Knowledge-based question answering}: GPQA-Diamond \citep{rein2024gpqa} evaluates factual knowledge and complex reasoning. We include it to assess whether math-centric training with different prior prompts can elicit behaviors associated with knowledge recall. While this ability may appear unrelated to solving math problems, it can be useful for recalling definitions, theorems, or formulas relevant to a given problem.
\end{itemize}

Additional metadata of these evaluation benchmarks, along with our training set, is available in \Cref{tab:dataset-overview}.

\begin{table*}[htbp]
  \centering
  \small
  \begin{tabular}{lccrcc}
    \toprule
    \textbf{Dataset}       & \textbf{Task} & \textbf{Split} & \textbf{Count} & \textbf{Answer Type} & \textbf{License} \\
    \midrule
    \multicolumn{6}{l}{\textbf{Training dataset}} \\
    STILLv3 \citep{chen2025empiricalstudyelicitingimproving}                & Math          & Train          & 29925         & N/A                  & N/A              \\
    \midrule
    \multicolumn{6}{l}{\textbf{Evaluation benchmark}} \\
    AIME24 \citep{li2024numinamath}                 & Math          & Test           & 30              & Number                 & N/A              \\
    AMC12 ’22–’23 \citep{li2024numinamath}          & Math          & Test           & 83              & Number                 & N/A              \\
    MATH-500 \citep{hendrycks2021measuring}               & Math          & Test           & 500             & Number                 & MIT License      \\
    HumanEval+ \citep{evalplus}             & Code          & Test           & 164             & Code                 & Apache 2.0       \\
    GPQA-Diamond \citep{rein2024gpqa}           & QA            & Test           & 198             & MC                   & CC BY 4.0        \\
    \bottomrule
  \end{tabular}
  \caption{Overview of the training dataset and evaluation benchmarks. We note that all datasets are available for the purposes used in this study.}
  \label{tab:dataset-overview}
\end{table*}

\subsubsection{Qualitative Analysis}\label{sec:qual_eval}

To investigate whether different pPE approaches lead to distinct behavioral patterns after RFT, we assess the following aspects:

\paragraph{Training dynamics and response length} 
We analyze whether different pPE approaches result in distinct training dynamics across models. In addition, we compute the average number of tokens in generated responses.

\paragraph{Four fundamental cognitive behaviors of reasoning models} 
\citet{gandhi2025cognitivebehaviorsenableselfimproving} identify four fundamental cognitive behaviors commonly exhibited by reasoning models. These behaviors are considered core components of what makes a model capable of reasoning: (1) \textbf{Verification}: Identifying errors in intermediate results, (2) \textbf{Backtracking}: Abandoning the current approach and trying alternatives, (3) \textbf{Subgoal setting}: Breaking problems down into smaller, more manageable steps, and (4) \textbf{Backward chaining}: Reasoning backward from the expected answer to the given inputs.

Following their methodology, we use an LLM-based classifier to detect the presence of each behavior in model outputs. While the original study used \texttt{gpt-4o-mini}, we employ a more recent model, \texttt{gpt-4.1-mini-2025-04-14}, for classification. Our goal is to compare the distribution of these behaviors across models trained or prompted using different iPE/pPE strategies. Prompts used for classification are provided in \Cref{fig:class_verification,fig:class_backtracking,fig:class_subgoal,fig:class_backward_chaining}.

\begin{figure}[htbp]
\footnotesize
\centering
\begin{tcolorbox}[title=Verifications Classification Prompt, colback=gray!5, colframe=amethyst!75!black]
\input{prompts/four_classification/verifications.txt}
\end{tcolorbox}
\caption{Prompt used to identify verification behavior--explicit checking or validation of intermediate results--in the model's reasoning.}
\label{fig:class_verification}
\end{figure}

\begin{figure}[htbp]
\footnotesize
\centering
\begin{tcolorbox}[title=Backtracking Classification Prompt, colback=gray!5, colframe=amethyst!75!black]
\input{prompts/four_classification/backtracking.txt}
\end{tcolorbox}
\caption{Classification prompt used to detect instances of backtracking--when a model revises or abandons a previous approach--in its reasoning trace.}
\label{fig:class_backtracking}
\end{figure}

\begin{figure}[htbp]
\footnotesize
\centering
\begin{tcolorbox}[title=Subgoal Settings Classification Prompt, colback=gray!5, colframe=amethyst!75!black]
\input{prompts/four_classification/subgoal_settings.txt}
\end{tcolorbox}
\caption{Prompt used to detect subgoal setting behavior, where the model breaks a problem into smaller, intermediate steps.}
\label{fig:class_subgoal}
\end{figure}

\begin{figure}[htbp]
\footnotesize
\centering
\begin{tcolorbox}[title=Backward Chaining Classification Prompt, colback=gray!5, colframe=amethyst!75!black]
\input{prompts/four_classification/backward_chaining.txt}
\end{tcolorbox}
\caption{Classification prompt used to identify backward chaining--reasoning from the goal back to known facts--within a model’s response.}
\label{fig:class_backward_chaining}
\end{figure}

\paragraph{Five pPE-specific behaviors}
We further adapt the same classification approach to evaluate whether the target behavior elicited by each pPE (e.g., reasoning, planning, coding, knowledge recall, or example generation) is present in model responses. This is treated as a binary classification task, assessing the presence or absence of the expected behavior per response. Classification prompts are provided in \Cref{fig:class_think,fig:class_plan,fig:class_code,fig:class_knowledge,fig:class_examples}.

\begin{figure}[htbp]
\footnotesize
\centering
\begin{tcolorbox}[title=Think Classification Prompt, colback=gray!5, colframe=amethyst!75!black]
\input{prompts/five_classification/think.txt}
\end{tcolorbox}
\caption{Prompt used to classify whether the model is exhibiting reasoning aligned with the \texttt{<think>} prompt (step-by-step logical reasoning).}
\label{fig:class_think}
\end{figure}

\begin{figure}[htbp]
\footnotesize
\centering
\begin{tcolorbox}[title=Plan Classification Prompt, colback=gray!5, colframe=amethyst!75!black]
\input{prompts/five_classification/plan.txt}
\end{tcolorbox}
\caption{Prompt used to identify whether the model is engaging in explicit planning, consistent with the \texttt{<plan>} prompting style.}
\label{fig:class_plan}
\end{figure}

\begin{figure}[htbp]
\footnotesize
\centering
\begin{tcolorbox}[title=Code Classification Prompt, colback=gray!5, colframe=amethyst!75!black]
\input{prompts/five_classification/code.txt}
\end{tcolorbox}
\caption{Prompt used to determine whether code-based reasoning patterns, encouraged by the \texttt{<code>} prompt, are present in the model output.}
\label{fig:class_code}
\end{figure}

\begin{figure}[htbp]
\footnotesize
\centering
\begin{tcolorbox}[title=Knowledge Classification Prompt, colback=gray!5, colframe=amethyst!75!black]
\input{prompts/five_classification/knowledge.txt}
\end{tcolorbox}
\caption{Prompt used to detect knowledge-recall behavior, as in \texttt{<knowledge>} approach.}
\label{fig:class_knowledge}
\end{figure}

\begin{figure}[htbp]
\footnotesize
\centering
\begin{tcolorbox}[title=Examples Classification Prompt, colback=gray!5, colframe=amethyst!75!black]
\input{prompts/five_classification/examples.txt}
\end{tcolorbox}
\caption{Prompt used to classify whether the model is generating illustrative examples, as intended by the \texttt{<examples>} prompting approach.}
\label{fig:class_examples}
\end{figure}

For all analyses, we exclude HumanEval+ due to missing raw responses from the evaluation program. We also omit a very small number of responses (less than 0.05\% of the total) due to response parsing errors from \texttt{gpt-4.1-mini-2025-04-14}.

\section{Additional Results}\label{sec:additional_results}

This section presents additional results from the generalization studies discussed in \Cref{sec:generalization}. Training dynamics for the three language models are provided in \Cref{sec:dynamics_response_len}. Results for the classification of the four fundamental behaviors and the five elicited behavior categories are available in \Cref{sec:four_behaviors} and \Cref{sec:five_behaviors}, respectively. In addition, we provide an alternative visualization of performance across benchmarks from the main experiments on Qwen2.5-7B, showing changes over the baseline in \Cref{tab:main-perf-change-colored}.

\begin{table}[htbp]
  \centering
  \tiny
  \begin{tabular}{lrrrrr|r}
      \toprule
      \textbf{Model} 
        & \textbf{AIME} 
        & \textbf{AMC} 
        & \textbf{GPQA} 
        & \textbf{MATH} 
        & \textbf{HE+} 
        & \textbf{Avg.} \\
      \midrule
      \multicolumn{7}{l}{\textbf{\underline{iPE}}} \\
      \quad Think                 
        & \cellcolor{neg}-3.33 
        & \cellcolor{neg}-6.02 
        & +0.00 
        & \cellcolor{pos}+0.40 
        & \cellcolor{pos}\underline{+2.40} 
        & \cellcolor{neg}-1.31 \\
      \quad Plan                  
        & \cellcolor{neg}-3.33 
        & \cellcolor{neg}-7.23 
        & +0.00 
        & \cellcolor{neg}-4.40 
        & \cellcolor{pos}+1.20 
        & \cellcolor{neg}-2.75 \\
      \quad Code                  
        & +0.00 
        & \cellcolor{neg}-10.84 
        & +0.00 
        & \cellcolor{neg}-4.20 
        & \cellcolor{neg}-0.60 
        & \cellcolor{neg}-3.13 \\
      \quad Knowledge             
        & \cellcolor{pos}\underline{+6.67} 
        & \cellcolor{neg}-12.05 
        & +0.00 
        & \cellcolor{pos}+4.00 
        & \cellcolor{neg}-0.60 
        & \cellcolor{neg}-0.40 \\
      \quad Examples              
        & \cellcolor{pos}+3.34 
        & \cellcolor{neg}-4.82
        & +0.00 
        & \cellcolor{pos}+1.20 
        & \cellcolor{neg}-72.60 
        & \cellcolor{neg}-14.58 \\
      \midrule
      \multicolumn{7}{l}{\textbf{\underline{RFT}}} \\
      \quad No PP   
        & \cellcolor{pos}\textbf{+13.34}
        & +0.00
        & \cellcolor{neg}-3.03
        & \cellcolor{pos}+14.80
        & \cellcolor{pos}+1.20
        & \cellcolor{pos}+4.79 \\
      \midrule
      \multicolumn{7}{l}{\textbf{\underline{pPE}}} \\
      \quad Think   
        & \cellcolor{pos}\underline{+6.67} 
        & \cellcolor{pos}+6.02 
        & \cellcolor{pos}\underline{+4.04} 
        & \cellcolor{pos}\textbf{+17.60} 
        & \cellcolor{neg}-2.50 
        & \cellcolor{pos}\underline{+6.37} \\
      \quad Plan    
        & \cellcolor{pos}\underline{+6.67} 
        & \cellcolor{pos}\underline{+7.23} 
        & \cellcolor{pos}+0.51 
        & \cellcolor{pos}+14.00 
        & \cellcolor{neg}-3.70 
        & \cellcolor{pos}+4.94 \\
      \quad Code    
        & \cellcolor{pos}+3.34 
        & \cellcolor{pos}\textbf{+9.64} 
        & \cellcolor{pos}+1.01 
        & \cellcolor{pos}+10.60 
        & \cellcolor{pos}\textbf{+5.40} 
        & \cellcolor{pos}+6.00 \\
      \quad Knowledge
        & \cellcolor{pos}+3.34 
        & +0.00 
        & \cellcolor{neg}-2.52 
        & \cellcolor{pos}+15.40 
        & \cellcolor{pos}+0.60 
        & \cellcolor{pos}+3.36 \\
      \quad Examples
        & \cellcolor{pos}\underline{+6.67} 
        & \cellcolor{pos}+6.02 
        & \cellcolor{pos}\textbf{+6.57} 
        & \cellcolor{pos}\underline{+15.60} 
        & +0.00 
        & \cellcolor{pos}\textbf{+6.97} \\
      \bottomrule
    \end{tabular}
  \caption{Absolute change in accuracy (green = gain, red = drop) relative to the zero-shot Qwen2.5-7B base model.}
  \label{tab:main-perf-change-colored}
\end{table}

\subsection{Training Dynamics and Average Response Length}\label{sec:dynamics_response_len}

\begin{figure}[htbp]
  \centering
  \includegraphics[width=\linewidth]{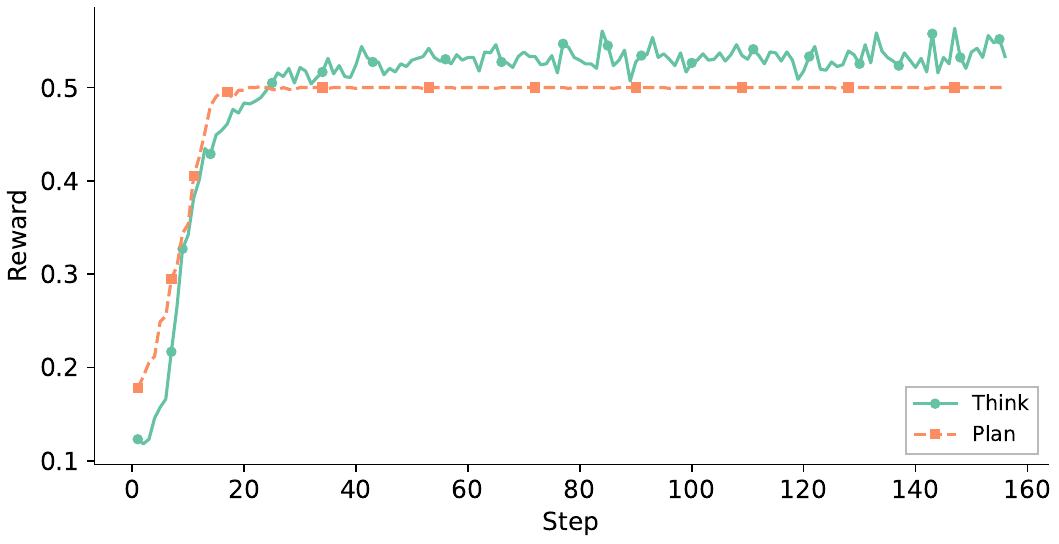}
  \caption{Reward progression for Qwen2.5-3B during RFT.}
  \label{fig:3b_reward}
\end{figure}

\begin{figure}[htbp]
  \centering
  \includegraphics[width=\linewidth]{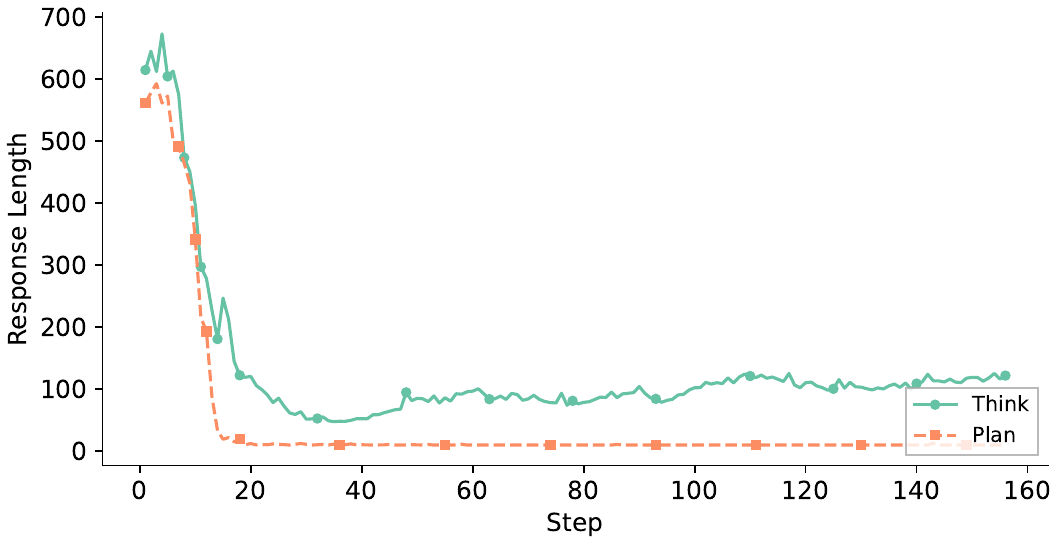}
  \caption{Evolution of the average response length for Qwen2.5-3B during RFT.}
  \label{fig:3b_response_length}
\end{figure}

\begin{figure}[htbp]
  \centering
  \includegraphics[width=\linewidth]{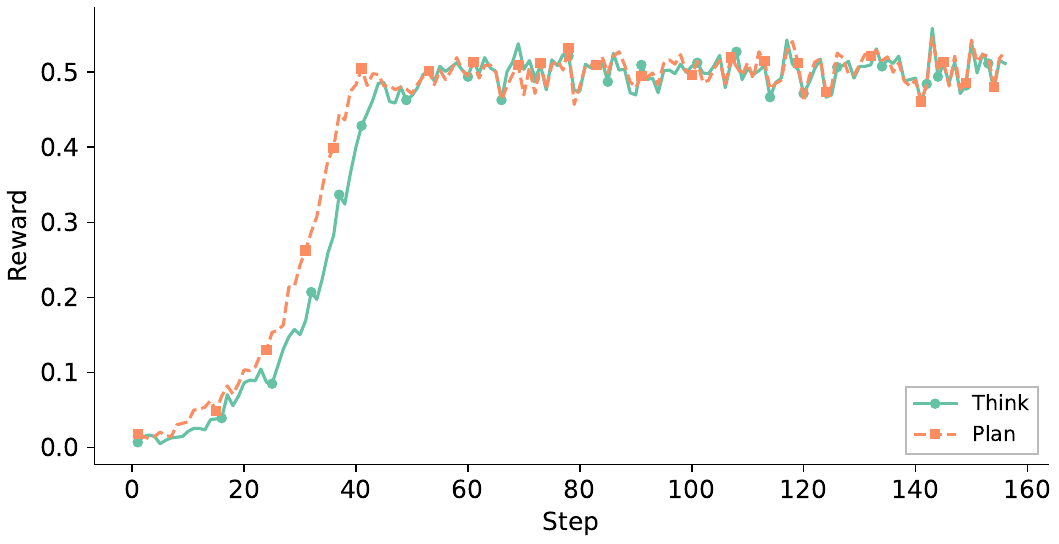}
  \caption{Reward progression for Llama 3.1-8B during RFT.}
  \label{fig:llama31_reward}
\end{figure}

\begin{figure}[htbp]
  \centering
  \includegraphics[width=\linewidth]{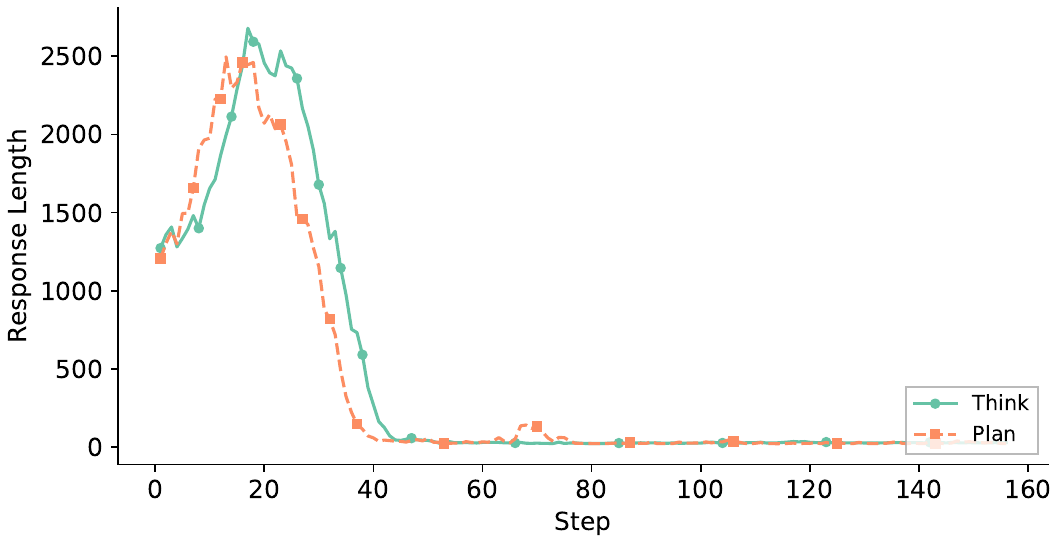}
  \caption{Evolution of the average response length for Llama 3.1-8B during RFT.}
  \label{fig:llama31_response_length}
\end{figure}

\begin{figure}[htbp]
  \centering
  \includegraphics[width=\linewidth]{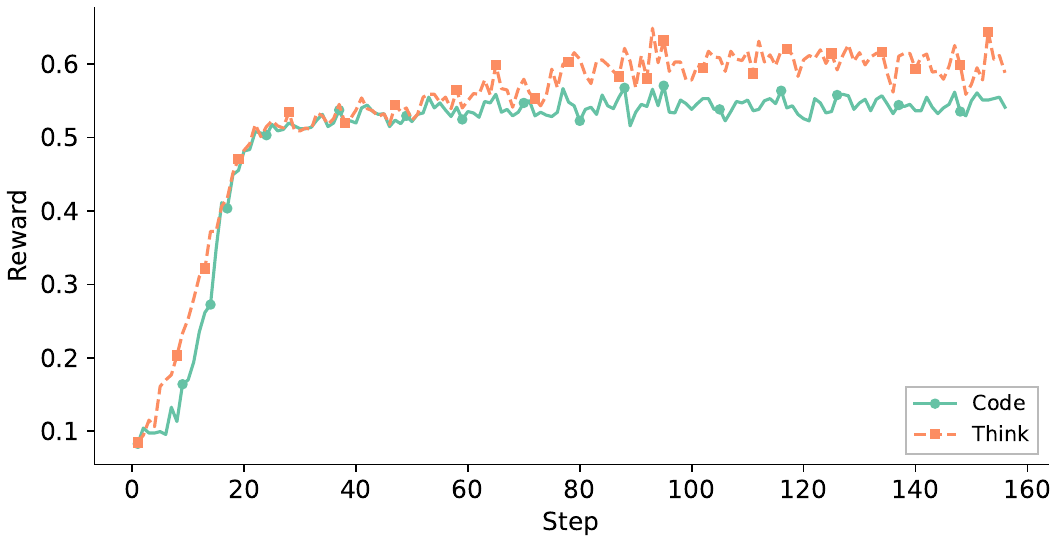}
  \caption{Reward progression for Qwen2.5-Coder-7B during RFT.}
  \label{fig:coder_reward}
\end{figure}

\begin{figure}[htbp]
  \centering
  \includegraphics[width=\linewidth]{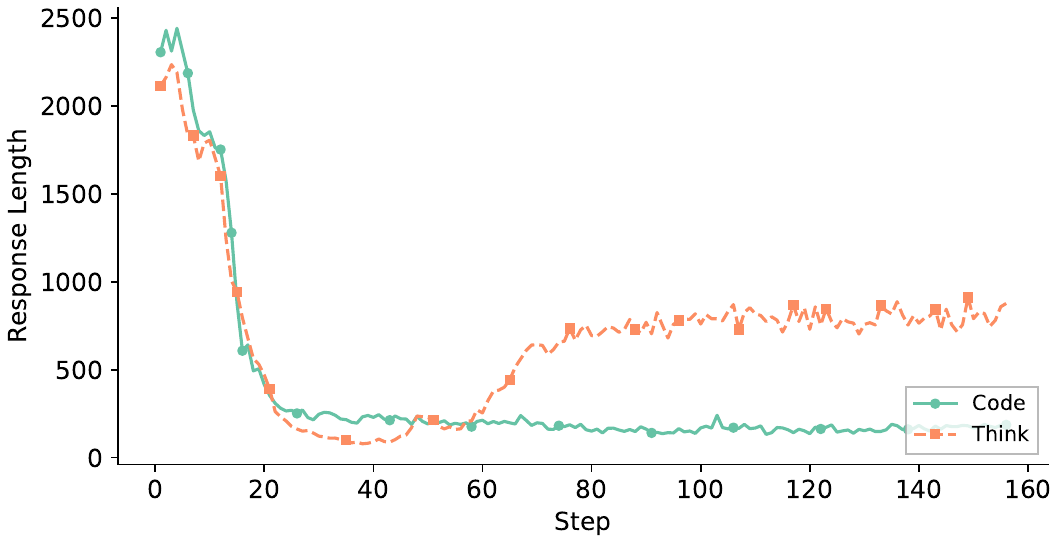}
  \caption{Evolution of the average response length for Qwen2.5-Coder-7B during RFT.}
  \label{fig:coder_response_length}
\end{figure}

\Cref{fig:3b_reward,fig:3b_response_length,fig:llama31_reward,fig:llama31_response_length,fig:coder_reward,fig:coder_response_length} present the reward and response-length dynamics during RFT for Qwen2.5-3B, Llama 3.1-8B, and Qwen2.5-Coder-7B, respectively.

\begin{table}[htbp]
  \centering
  \tiny
  \setlength\doublerulesep{0.5pt}
  \setlength{\tabcolsep}{5pt}
  \begin{tabular}{lrrrrr}
    \toprule
    \textbf{Model}
      & \textbf{AIME}
      & \textbf{AMC}
      & \textbf{GPQA}
      & \textbf{MATH}
      & \textbf{Avg.} \\
    \midrule
    \textbf{Qwen2.5 3B}
      & 1249.00
      & 1297.70
      & 455.85
      & 981.93
      & 996.12 \\
    \midrule
    \multicolumn{6}{l}{\textbf{\underline{iPE}}} \\
    \quad Think
      & 1764.63
      & 1156.81
      & 455.85
      & 754.45
      & 1032.94 \\
    \quad Plan
      & 1349.13
      & 1270.47
      & 455.85
      & 631.23
      & 926.67 \\
    \midrule
    \multicolumn{6}{l}{\textbf{\underline{pPE}}} \\
    \quad Think
      & 124.37
      & 492.70
      & 412.17
      & 182.44
      & 302.92 \\
    \quad Plan
      & 9.00
      & 9.00
      & 274.95
      & 9.00
      & 75.49 \\
    \midrule\midrule
    \textbf{Llama 3.1-8B}
      & 8.03
      & 11.25
      & N/A
      & 128.93
      & 49.40 \\
    \midrule
    \multicolumn{6}{l}{\textbf{\underline{iPE}}} \\
    \quad Think
      & 6122.33
      & 4347.72
      & N/A
      & 4177.73
      & 4882.59 \\
    \quad Plan
      & 4562.97
      & 5150.96
      & N/A
      & 4452.76
      & 4722.23 \\
    \midrule
    \multicolumn{6}{l}{\textbf{\underline{pPE}}} \\
    \quad Think
      & 22.70
      & 23.23
      & N/A
      & 25.84
      & 23.92 \\
    \quad Plan
      & 21.10
      & 20.37
      & N/A
      & 53.94
      & 31.80 \\
    \midrule\midrule
    \textbf{Qwen2.5-Coder-7B}
      & 3856.83
      & 2947.86
      & 1510.95
      & 3570.15
      & 2971.45 \\
    \midrule
    \multicolumn{6}{l}{\textbf{\underline{iPE}}} \\
    \quad Think
      & 2601.03
      & 748.31
      & 1510.95
      & 917.51
      & 1444.45 \\
    \quad Code
      & 779.67
      & 337.59
      & 1510.95
      & 265.43
      & 723.41 \\
    \midrule
    \multicolumn{6}{l}{\textbf{\underline{pPE}}} \\
    \quad Think
      & 1850.67
      & 1045.12
      & 592.05
      & 758.67
      & 1061.63 \\
    \quad Code
      & 164.72
      & 254.41
      & 311.49
      & 152.67
      & 220.82 \\
    \bottomrule
  \end{tabular}
  \caption{Average response length, i.e., number of tokens, of Qwen2.5-3B, Llama 3.1-8B, and Qwen2.5-Coder-7B when prompted with different iPE or RFT with different pPE approaches across four benchmarks.}
  \label{tab:generalization-response-length}
\end{table}

Finally, \Cref{tab:generalization-response-length} reports the average response length, i.e., average number of tokens in responses for the generalization experiments.

\subsection{Four Fundamental Cognitive Behaviors}\label{sec:four_behaviors}

\begin{figure}[htbp]
  \centering
  \includegraphics[width=\linewidth]{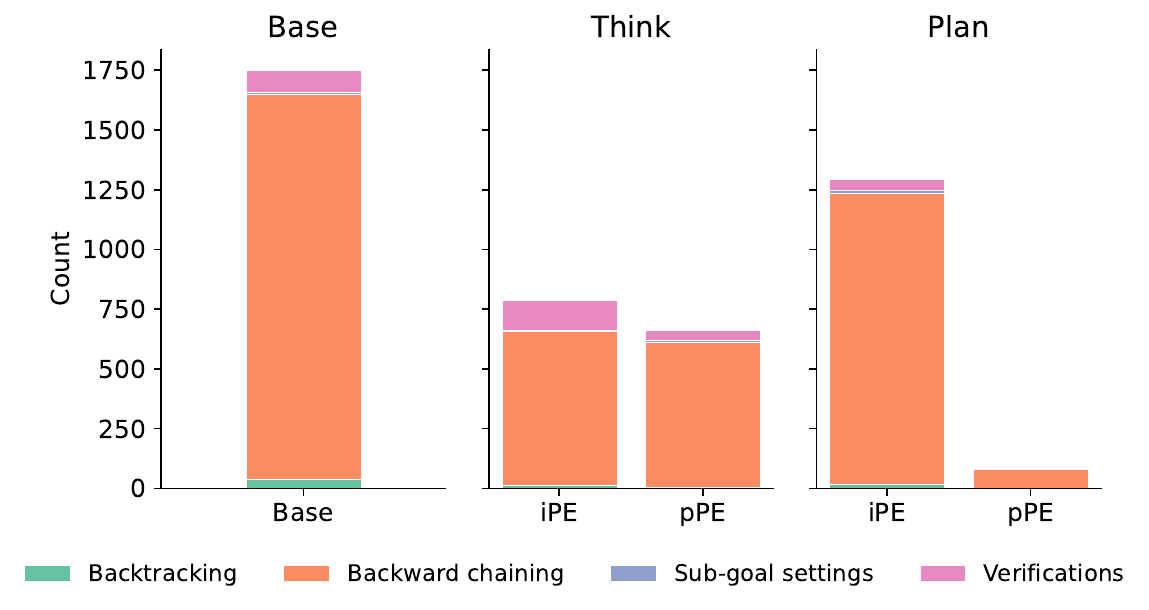}
  \caption{Ratio of the four fundamental cognitive behaviors--backtracking, backward chaining, subgoal setting, and verification--across different prompting (iPE) and RFT (pPE) approaches with Qwen2.5-3B.}
  \label{fig:3b_four_behav}
\end{figure}

\begin{figure}[htbp]
  \centering
  \includegraphics[width=\linewidth]{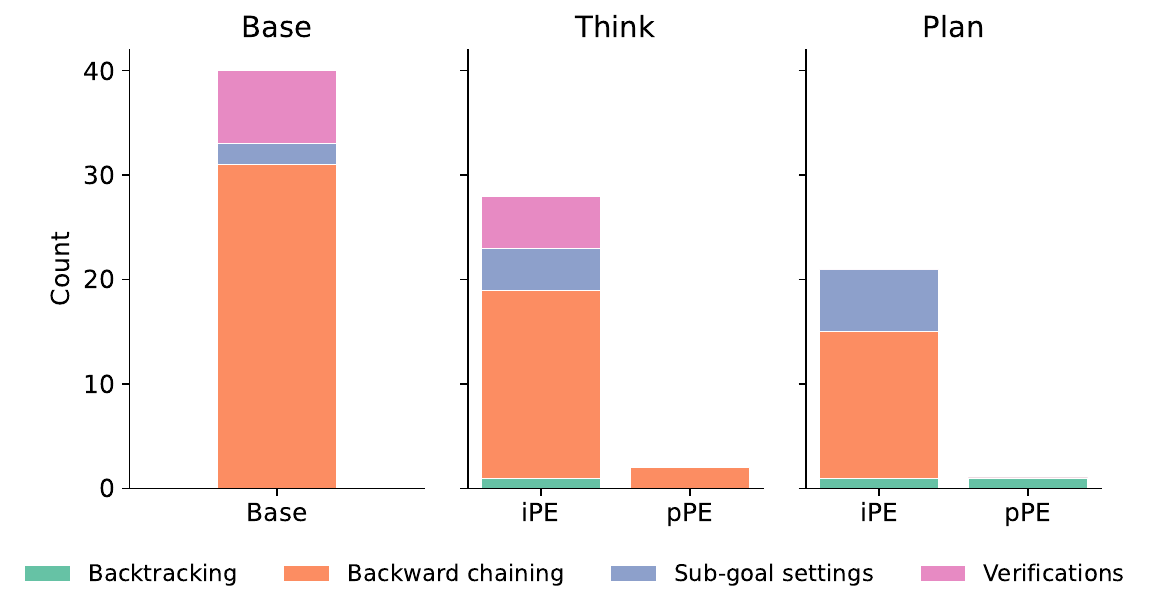}
  \caption{Ratio of the four fundamental cognitive behaviors--backtracking, backward chaining, subgoal setting, and verification--across different prompting (iPE) and RFT (pPE) approaches with Llama 3.1-8B.}
  \label{fig:llama_four_behav}
\end{figure}

\begin{figure}[htbp]
  \centering
  \includegraphics[width=\linewidth]{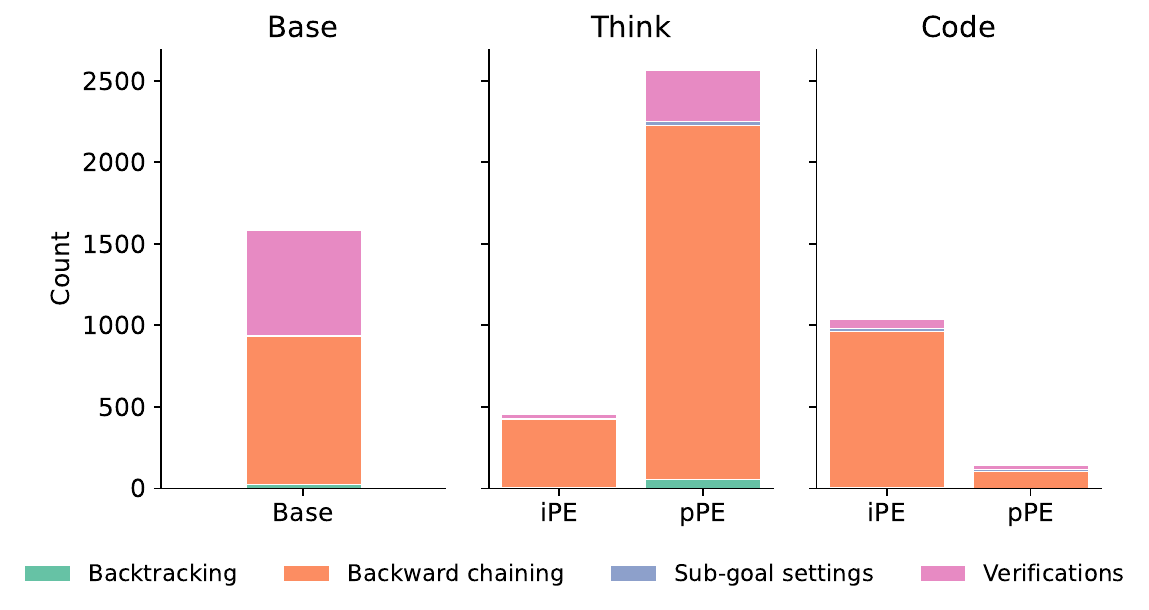}
  \caption{Ratio of the four fundamental cognitive behaviors--backtracking, backward chaining, subgoal setting, and verification--across different prompting (iPE) and RFT (pPE) approaches with Qwen2.5-Coder-7B.}
  \label{fig:coder_four_behav}
\end{figure}

In this subsection, we present the results of four fundamental cognitive behavior classifications from the generalization studies for Qwen2.5-3B, Llama 3.1-8B, and Qwen2.5-Coder-7B, shown in \Cref{fig:3b_four_behav,fig:llama_four_behav,fig:coder_four_behav}, respectively.

In addition, we provide detailed tables of the exact behavior counts for the main experiment in \Cref{tab:main-four-behavior-counts}, previously visualized in \Cref{fig:main_four_behav}, and for the generalization studies in \Cref{tab:generalization-four-behavior-counts}, previously visualized in \Cref{fig:3b_four_behav,fig:llama_four_behav,fig:coder_four_behav}.

\begin{table}[htbp]
  \centering
  \tiny
  \begin{tabular}{lrrrr}
    \toprule
    \textbf{Model}
      & \textbf{Backtrack.}
      & \textbf{Back.Chain.}
      & \textbf{Subgoal.Set.}
      & \textbf{Veri.} \\
    \midrule
    \textbf{Qwen2.5-7B}
      & 30
      & 1707
      & 9
      & 195 \\
    \midrule
    \multicolumn{5}{l}{\textbf{iPE}} \\
    \quad Think
      & 31
      & 1542
      & 15
      & 201 \\
    \quad Plan
      & 60
      & 2202
      & \underline{6}
      & 156 \\
    \quad Code
      & 66
      & 2394
      & 31
      & 223 \\
    \quad Knowledge
      & 79
      & 2794
      & \textbf{56}
      & 274 \\
    \quad Examples
      & \textbf{114}
      & \textbf{3159}
      & 30
      & \textbf{453} \\
    \midrule
    \multicolumn{5}{l}{\textbf{RFT}} \\
    \quad No PP
      & 70
      & 2195
      & 23
      & 336 \\
    \midrule
    \multicolumn{5}{l}{\textbf{pPE}} \\
    \quad Think
      & \underline{27}
      & 1354
      & 16
      & 150 \\
    \quad Plan
      & 32
      & 2450
      & 11
      & 110 \\
    \quad Code
      & 42
      & 1628
      & 19
      & 262 \\
    \quad Knowledge
      & 63
      & \underline{1180}
      & 14
      & 148 \\
    \quad Examples
      & 46
      & 1545
      & 8
      & \underline{98} \\
    \bottomrule
  \end{tabular}
  \caption{Occurrence counts of the four fundamental cognitive behaviors--backtracking (Backtrack.), backward chaining (Back.Chain.), subgoal settings (Subgoal.Set.), and verifications (Veri.)--in Qwen2.5-7B's responses under different iPE and pPE approaches. Counts are obtained via the LM-based classification framework of \citet{gandhi2025cognitivebehaviorsenableselfimproving}. \textbf{Bold} marks the highest count and \underline{underlined} marks the lowest count per column.}
  \label{tab:main-four-behavior-counts}
\end{table}

\begin{table}[htbp]
  \centering
  \tiny
  \setlength\doublerulesep{0.5pt}
  \setlength{\tabcolsep}{5pt}
  \begin{tabular}{lrrrr}
    \toprule
    \textbf{Model}
      & \textbf{Backtrack.}
      & \textbf{Back.Chain.}
      & \textbf{Subgoal.Set.}
      & \textbf{Veri.} \\
    \midrule
    \textbf{Qwen2.5-3B}
      & \textbf{37}
      & \textbf{1611}
      & 10
      & 91 \\
    \midrule
    \multicolumn{5}{l}{\textbf{iPE}} \\
    \quad Think
      & 13
      & 646
      & 5
      & \textbf{125} \\
    \quad Plan
      & 16
      & 1220
      & \textbf{13}
      & 46 \\
    \midrule
    \multicolumn{5}{l}{\textbf{pPE}} \\
    \quad Think
      & 6
      & 604
      & 12
      & 42 \\
    \quad Plan
      & \underline{0}
      & \underline{78}
      & \underline{0}
      & \underline{6} \\
    \midrule\midrule
    \textbf{Llama 3.1-8B}
      & 0
      & \textbf{31}
      & 2
      & \textbf{7} \\
    \midrule
    \multicolumn{5}{l}{\textbf{iPE}} \\
    \quad Think
      & \textbf{1}
      & 18
      & 4
      & 5 \\
    \quad Plan
      & \textbf{1}
      & 14
      & \textbf{6}
      & \underline{0} \\
    \midrule
    \multicolumn{5}{l}{\textbf{pPE}} \\
    \quad Think
      & \underline{0}
      & 2
      & \underline{0}
      & \underline{0} \\
    \quad Plan
      & \textbf{1}
      & \underline{0}
      & \underline{0}
      & \underline{0} \\
    \midrule\midrule
    \textbf{Qwen2.5-Coder-7B}
      & 25
      & 908
      & 7
      & \textbf{644} \\
    \midrule
    \multicolumn{5}{l}{\textbf{iPE}} \\
    \quad Think
      & \underline{6}
      & 420
      & \underline{3}
      & \underline{24} \\
    \quad Code
      & 6
      & 959
      & 17
      & 53 \\
    \midrule
    \multicolumn{5}{l}{\textbf{pPE}} \\
    \quad Think
      & \textbf{57}
      & \textbf{2172}
      & \textbf{21}
      & 313 \\
    \quad Code
      & \underline{2}
      & \underline{104}
      & 12
      & 26 \\
    \bottomrule
  \end{tabular}
  \caption{Occurrence counts of the four fundamental cognitive behaviors--backtracking (Backtrack.), backward chaining (Back.Chain.), subgoal settings (Subgoal.Set.), and verifications (Veri.)--in Qwen2.5-3B, Llama 3.1-8B, and Qwen2.5-Coder-7B responses under different iPE and pPE approaches. Counts are obtained via the LM-based classification framework of \citet{gandhi2025cognitivebehaviorsenableselfimproving}. \textbf{Bold} marks the highest count and \underline{underlined} marks the lowest count per column under the same base model.}
  \label{tab:generalization-four-behavior-counts}
\end{table}

\subsection{Five Elicited Behaviors}\label{sec:five_behaviors}

\begin{figure}[htbp]
  \centering
  \includegraphics[width=\linewidth]{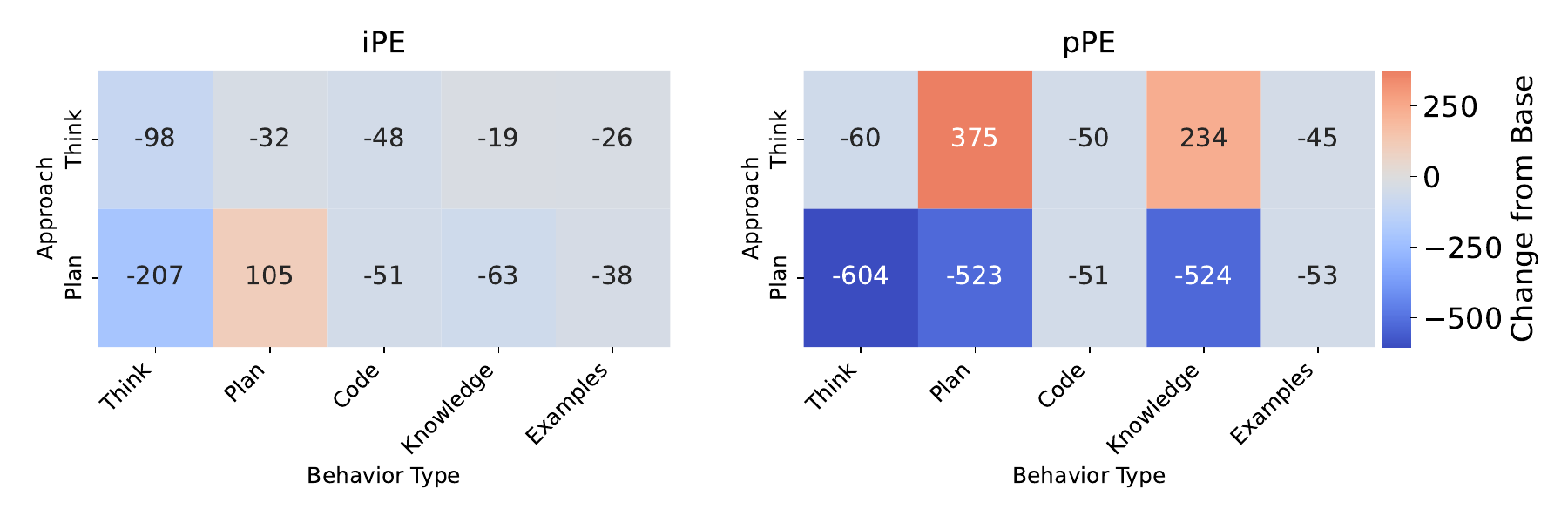}
  \caption{Behavior alignment heatmaps for Qwen2.5 3B: iPE on the left, RFT on the right.}
  \label{fig:3b_five_template}
\end{figure}

\begin{figure}[htbp]
  \centering
  \includegraphics[width=\linewidth]{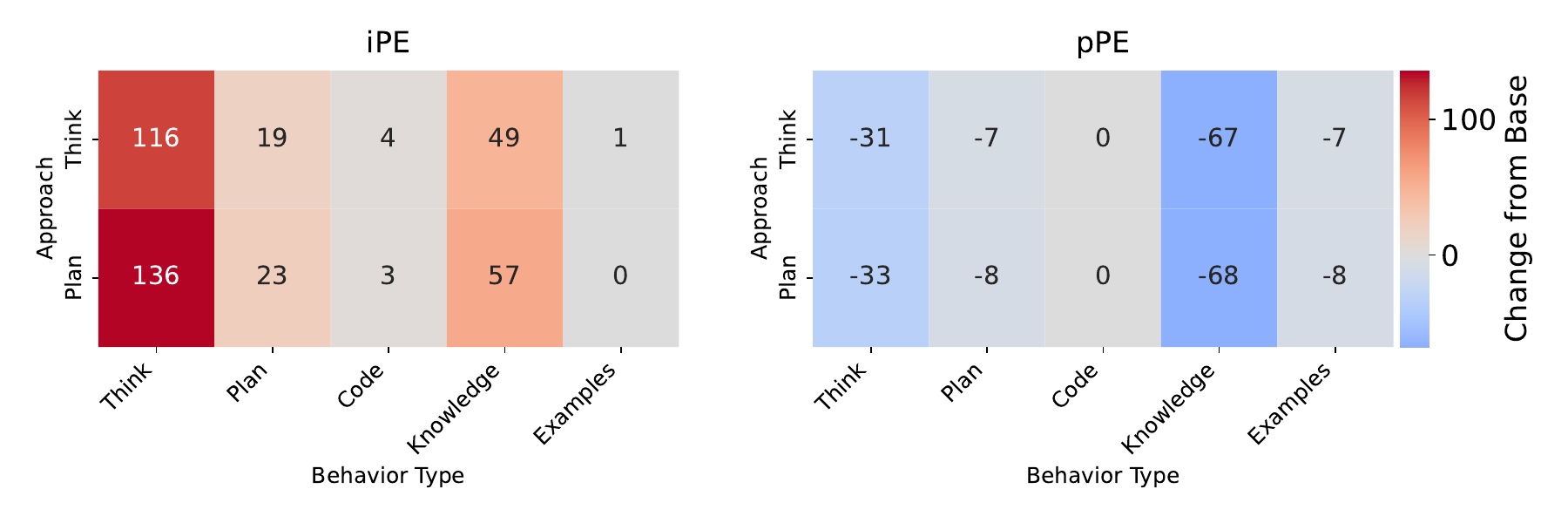}
  \caption{Behavior activation patterns for Llama 3.1 8B under iPE (left) and RFT (right).}
  \label{fig:llama_five_template}
\end{figure}

\begin{figure}[htbp]
  \centering
  \includegraphics[width=\linewidth]{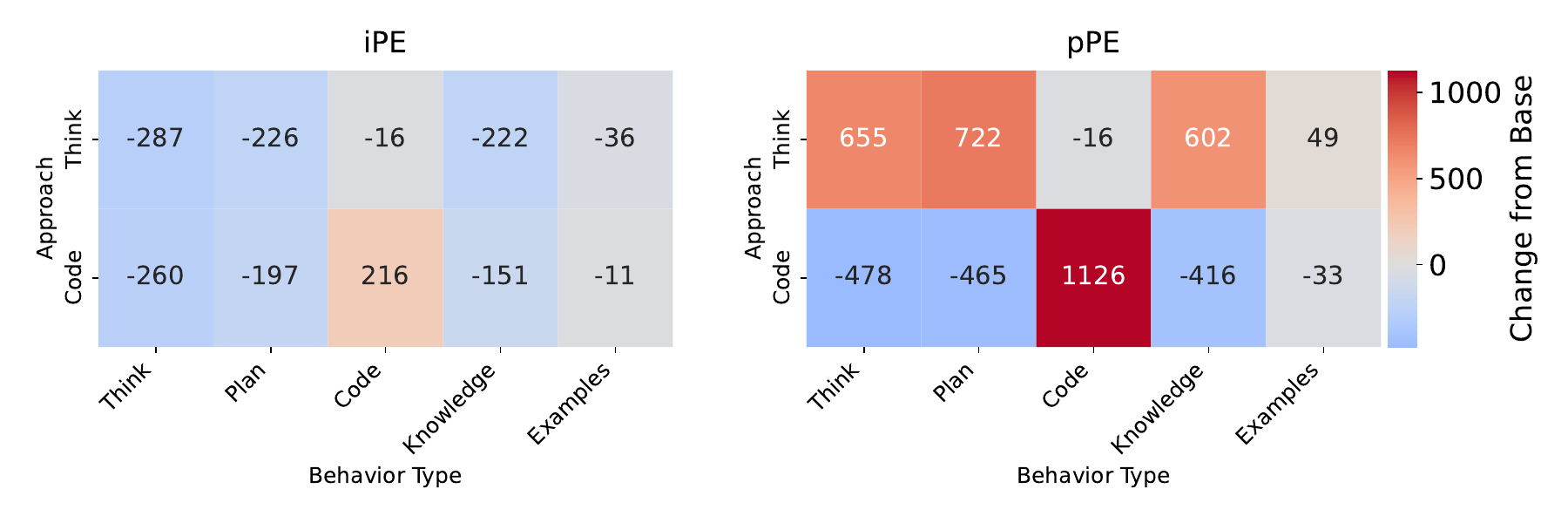}
  \caption{Behavior heatmaps for Qwen2.5-Coder 7B: iPE (left) vs.\ RFT (right).}
  \label{fig:coder_five_template}
\end{figure}

In this subsection, we present the results of five elicited cognitive behavior classifications from the generalization studies for Qwen2.5-3B, Llama 3.1-8B, and Qwen2.5-Coder-7B, shown in \Cref{fig:3b_five_template,fig:llama_five_template,fig:coder_five_template}, respectively.

In addition, we provide detailed tables of the exact behavior counts for the main experiment in \Cref{tab:main-five-behavior-counts}, previously visualized in \Cref{fig:main_five_template}, and for the generalization studies in \Cref{tab:generalization-five-behavior-counts}, previously visualized in \Cref{fig:3b_five_template,fig:llama_five_template,fig:coder_five_template}.

\begin{table}[htbp]
  \centering
  \tiny
  \begin{tabular}{lrrrrr}
    \toprule
    \textbf{Model}
      & \textbf{Code}
      & \textbf{Examples}
      & \textbf{Knowledge}
      & \textbf{Plan}
      & \textbf{Think} \\
    \midrule
    \textbf{Qwen2.5-7B}
      & 193
      & 77
      & 626
      & 652
      & 711 \\
    \midrule
    \multicolumn{6}{l}{\textbf{iPE}} \\
    \quad Think
      & 31
      & 55
      & 635
      & \underline{622}
      & 716 \\
    \quad Plan
      & 38
      & \underline{51}
      & \underline{616}
      & 733
      & \underline{659} \\
    \quad Code
      & \textbf{704}
      & 60
      & 1002
      & 1137
      & 1130 \\
    \quad Knowledge
      & 88
      & 112
      & \textbf{1168}
      & \textbf{1155}
      & 1289 \\
    \quad Examples
      & 68
      & \textbf{159}
      & 1112
      & 1136
      & \textbf{1304} \\
    \midrule
    \multicolumn{6}{l}{\textbf{RFT}} \\
    \quad No PP
      & 16
      & 75
      & 708
      & 716
      & 780 \\
    \midrule
    \multicolumn{6}{l}{\textbf{pPE}} \\
    \quad Think
      & 34
      & 68
      & 662
      & 741
      & 764 \\
    \quad Plan
      & 23
      & 57
      & 658
      & 759
      & 757 \\
    \quad Code
      & 602
      & 58
      & 640
      & 688
      & 739 \\
    \quad Knowledge
      & \underline{14}
      & 61
      & 720
      & 722
      & 760 \\
    \quad Examples
      & 21
      & 57
      & 632
      & 704
      & 746 \\
    \bottomrule
  \end{tabular}
  \caption{Occurrence counts of the five elicited behaviors--code, examples, knowledge, plan, and think--in Qwen2.5-7B's responses under different iPE and pPE approaches. \textbf{Bold} marks the highest count and \underline{underlined} marks the lowest count per column.}
  \label{tab:main-five-behavior-counts}
\end{table}

\begin{table}[htbp]
  \centering
  \tiny
  \setlength\doublerulesep{0.5pt}
  \setlength{\tabcolsep}{5pt}
  \begin{tabular}{lrrrrr}
    \toprule
    \textbf{Model}
      & \textbf{Code}
      & \textbf{Examples}
      & \textbf{Knowledge}
      & \textbf{Plan}
      & \textbf{Think} \\
    \midrule
    \textbf{Qwen2.5-3B}
      & \textbf{51}
      & \textbf{53}
      & \underline{554}
      & 545
      & \textbf{630} \\
    \midrule
    \multicolumn{6}{l}{\textbf{iPE}} \\
    \quad Think
      & \underline{3}
      & \underline{27}
      & 535
      & 513
      & 532 \\
    \quad Plan
      & 0
      & 15
      & \underline{491}
      & \underline{650}
      & 423 \\
    \midrule
    \multicolumn{6}{l}{\textbf{pPE}} \\
    \quad Think
      & 1
      & 8
      & \textbf{788}
      & \textbf{920}
      & 570 \\
    \quad Plan
      & 0
      & 0
      & 30
      & 22
      & 26 \\
    \midrule\midrule
    \textbf{Llama 3.1-8B}
      & 0
      & \underline{8}
      & 71
      & 9
      & 36 \\
    \midrule
    \multicolumn{6}{l}{\textbf{iPE}} \\
    \quad Think
      & \textbf{4}
      & \textbf{9}
      & \underline{120}
      & \underline{28}
      & \underline{152} \\
    \quad Plan
      & \underline{3}
      & \underline{8}
      & \textbf{128}
      & \textbf{32}
      & \textbf{172} \\
    \midrule
    \multicolumn{6}{l}{\textbf{pPE}} \\
    \quad Think
      & 0
      & 1
      & 4
      & 2
      & 5 \\
    \quad Plan
      & 0
      & 0
      & 3
      & 1
      & 3 \\
    \midrule\midrule
    \textbf{Qwen2.5-Coder-7B}
      & 16
      & \underline{54}
      & \underline{577}
      & \underline{525}
      & \underline{669} \\
    \midrule
    \multicolumn{6}{l}{\textbf{iPE}} \\
    \quad Think
      & 0
      & 18
      & 355
      & 299
      & 382 \\
    \quad Code
      & \underline{232}
      & 43
      & 426
      & 328
      & 409 \\
    \midrule
    \multicolumn{6}{l}{\textbf{pPE}} \\
    \quad Think
      & 0
      & 103
      & \textbf{1179}
      & \textbf{1247}
      & \textbf{1324} \\
    \quad Code
      & \textbf{1142}
      & 21
      & 161
      & 60
      & 191 \\
    \bottomrule
  \end{tabular}
  \caption{Occurrence counts of the five elicited behaviors--code, examples, knowledge, plan, and think--in Qwen2.5-3B, Llama 3.1-8B, and Qwen2.5-Coder-7B responses under different iPE and pPE approaches. \textbf{Bold} indicates the best performance per column under the same base model; \underline{underlined} indicates the second best per column under the same base model.}
  \label{tab:generalization-five-behavior-counts}
\end{table}

\section{Qualitative Examples}\label{sec:qual_examples}

To illustrate the behavioral differences induced by each pPE approach, we present qualitative outputs from models trained with different prior prompts in response to a shared math problem. Examples of model responses for each pPE approach are shown in \Cref{fig:qual_think,fig:qual_plan,fig:qual_code,fig:qual_knowledge,fig:qual_examples}, using the shared prompt in \Cref{fig:qual_prompt}.

\begin{figure}[htbp]
\footnotesize
\centering
\begin{tcolorbox}[title=Example Prompt, colback=gray!5, colframe=amethyst!75!black]
\input{examples/prompt.txt}
\end{tcolorbox}
\caption{The shared prompt used to probe all models in this qualitative comparison. The question requires the model to determine the number of positive whole-number divisors of 196.}
\label{fig:qual_prompt}
\end{figure}

\begin{figure}[htbp]
\footnotesize
\centering
\begin{tcolorbox}[title=Qwen2.5 7B RFT with Think, colback=gray!5, colframe=amethyst!75!black]
\input{examples/think.txt}
\end{tcolorbox}
\caption{Response from Qwen2.5 7B trained with the \texttt{<think>} prior prompt. The model demonstrates step-by-step reasoning throughout its solution.}
\label{fig:qual_think}
\end{figure}

\begin{figure}[htbp]
\footnotesize
\centering
\begin{tcolorbox}[title=Qwen2.5 7B RFT with Plan, colback=gray!5, colframe=amethyst!75!black]
\input{examples/plan.txt}
\end{tcolorbox}
\caption{Response from Qwen2.5 7B trained with the \texttt{<plan>} prior prompt. The model outlines a structured plan before proceeding to execution.}
\label{fig:qual_plan}
\end{figure}

\begin{figure}[htbp]
\footnotesize
\centering
\begin{tcolorbox}[title=Qwen2.5 Coder 7B RFT with Code, colback=gray!5, colframe=amethyst!75!black]
\input{examples/code.txt}
\end{tcolorbox}
\caption{Response from Qwen2.5-Coder 7B trained with the \texttt{<code>} prior prompt. The model uses Python code to assist in its reasoning process.}
\label{fig:qual_code}
\end{figure}

\begin{figure}[htbp]
\footnotesize
\centering
\begin{tcolorbox}[title=Qwen2.5 7B RFT with Knowledge, colback=gray!5, colframe=amethyst!75!black]
\input{examples/knowledge.txt}
\end{tcolorbox}
\caption{Response from Qwen2.5 7B trained with the \texttt{<knowledge>} prior prompt. The model first recalls definitions and relevant facts before solving the problem.}
\label{fig:qual_knowledge}
\end{figure}

\begin{figure}[htbp]
\scriptsize
\centering
\begin{tcolorbox}[title=Qwen2.5 7B RFT with Examples, colback=gray!5, colframe=amethyst!75!black]
\input{examples/examples.txt}
\end{tcolorbox}
\caption{Response from Qwen2.5 7B trained with the \texttt{<examples>} prior prompt. The model introduces illustrative examples to support its reasoning.}
\label{fig:qual_examples}
\end{figure}

\section{Declaration of AI Assistance}\label{sec:ai_assistance}

We utilized ChatGPT for grammatical checking and LaTeX support of the content presented in this study but did not use it for the initial draft of this study. Cursor was utilized for trivial and boilerplate code completion during data analysis. We declare that all content presented and code utilized in this study has been reviewed and edited by the authors.

\end{document}